\documentclass[letterpaper]{article}
\usepackage[preprint]{arxiv_tsumego}
\usepackage[hyphens]{url}
\usepackage{graphicx}
\usepackage{natbib}
 \newcommand{\stmax}[1]{#1\rlap{${}^{\ast}$}}   
\usepackage{caption}
\usepackage{array}
\usepackage{booktabs}
\usepackage{amsmath}          
\newcommand{\eB}[1]{\underline{\mathbf{#1}}} 
\newcommand{\mB}[1]{\underline{#1}}     
\newcommand{\hB}[1]{\mathbf{#1}}        
\newcommand{\pairS}[2]{\makebox[2.05em][r]{$#1$}\hspace{0.18em}|\hspace{0.18em}\makebox[2.05em][l]{$#2$}} 
\usepackage{booktabs}
\usepackage{amsfonts}
\usepackage{amsmath,amssymb}
\usepackage{multirow}
\usepackage{subcaption}
\usepackage{enumitem}
\usepackage{xcolor}
\usepackage{algorithm}
\usepackage{algorithmic}

\newcommand{\tsumego}{\textsc{TsuGO}}
\newcommand{\swr}{\mathrm{SWR}}
\newcommand{\sfh}{\mathrm{SFH}}
\newcommand{\scr}{\mathrm{SCR}}
\newcommand{\ssc}{\mathrm{SSC}}
\newcommand{\sbc}{\mathrm{SBC}}
\newcommand{\itt}{\mathrm{ITT}}
\newcommand{\ipn}{\mathrm{IPN}}
\newcommand{\tmd}{\mathrm{TMD}}
\newcommand{\tnc}{\mathrm{TNC}}
\newcommand{\tmf}{\mathrm{TMF}}

\title{TsuGO: Probing Search Efficiency in LLM Reasoning via Go Life-and-Death Problems}
\author{
    Shunwen Bai\textsuperscript{\rm 1,2},
    Ziping Ma\textsuperscript{\rm 2},
    Chaoyang Zhang\textsuperscript{\rm 2,3},
    Yarong Wang\textsuperscript{\rm 1,2},
    Jiale Liu\textsuperscript{\rm 1},
    Zhen Qin\textsuperscript{\rm 1}\textsuperscript{\dag},
    Qingpei Guo\textsuperscript{\rm 2}\textsuperscript{\dag}
}
\affiliations{
    \textsuperscript{\rm 1}Zhejiang University \quad
    \textsuperscript{\rm 2}Ant Group \quad
    \textsuperscript{\rm 3}Central South University\\
    \texttt{\{shunwenbai,zhenqin\}@zju.edu.cn, qingpei.gqp@antgroup.com}
}

\begin{document}

\maketitle
\begingroup
\renewcommand{\thefootnote}{}
\footnotetext{This work was supported by Ant Group Research Intern Program.}
\renewcommand{\thefootnote}{\dag}
\footnotetext{Corresponding author.}
\endgroup

\begin{abstract}
The evaluation of LLM reasoning is moving from final-answer accuracy to process-level assessment, yet existing methods still fail to capture how models plan reasoning paths and allocate reasoning resources---that is, how they organize search.
Prior process-level methods focus on the coherence and redundancy of chain-of-thought (CoT), and most benchmark tasks have a single objective solvable by static capabilities such as derivation and tool use, leaving search organization unmeasured.
We introduce \tsumego{}, a process-level reasoning benchmark for evaluating Search Efficiency in LLM reasoning through Go life-and-death problems.
These problems provide closed and verifiable solution spaces with an inherent adversarial structure, making candidate generation, response checking, branch comparison, and backtracking necessary parts of reasoning rather than incidental trace patterns.
By constraining the solution space, \tsumego{} disentangles domain knowledge from search organization, parses CoT into a structured search tree, and reports Search Efficiency together with Token Efficiency and other diagnostic metrics and visualizations.
Experiments show that current LLMs remain far from stable tsumego solving: stronger models succeed by finding the correct candidate earlier and sustaining effort on productive branches, but most models still behave much closer to unguided search algorithms than to neural-guided KataGo.
Longer CoT or higher Token Efficiency does not necessarily imply better search.
Our results identify search organization and reasoning-resource allocation as missing dimensions in LLM reasoning evaluation.
\end{abstract}

\section{Introduction}
\label{sec:intro}

\begin{figure}[t]
  \centering
  \includegraphics[width=\columnwidth]{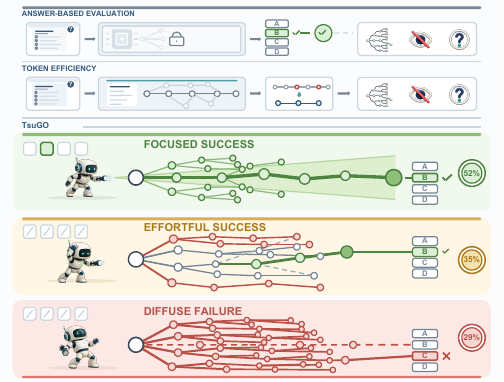}
  \vspace{-0.25em} 
  \caption{\tsumego{} reveals search organization that is hidden from Answer- and Token-Efficiency-based evaluation.}
  \label{fig:teaser}
\end{figure}

\begin{figure*}[t]
  \centering
  \includegraphics[width=\textwidth]{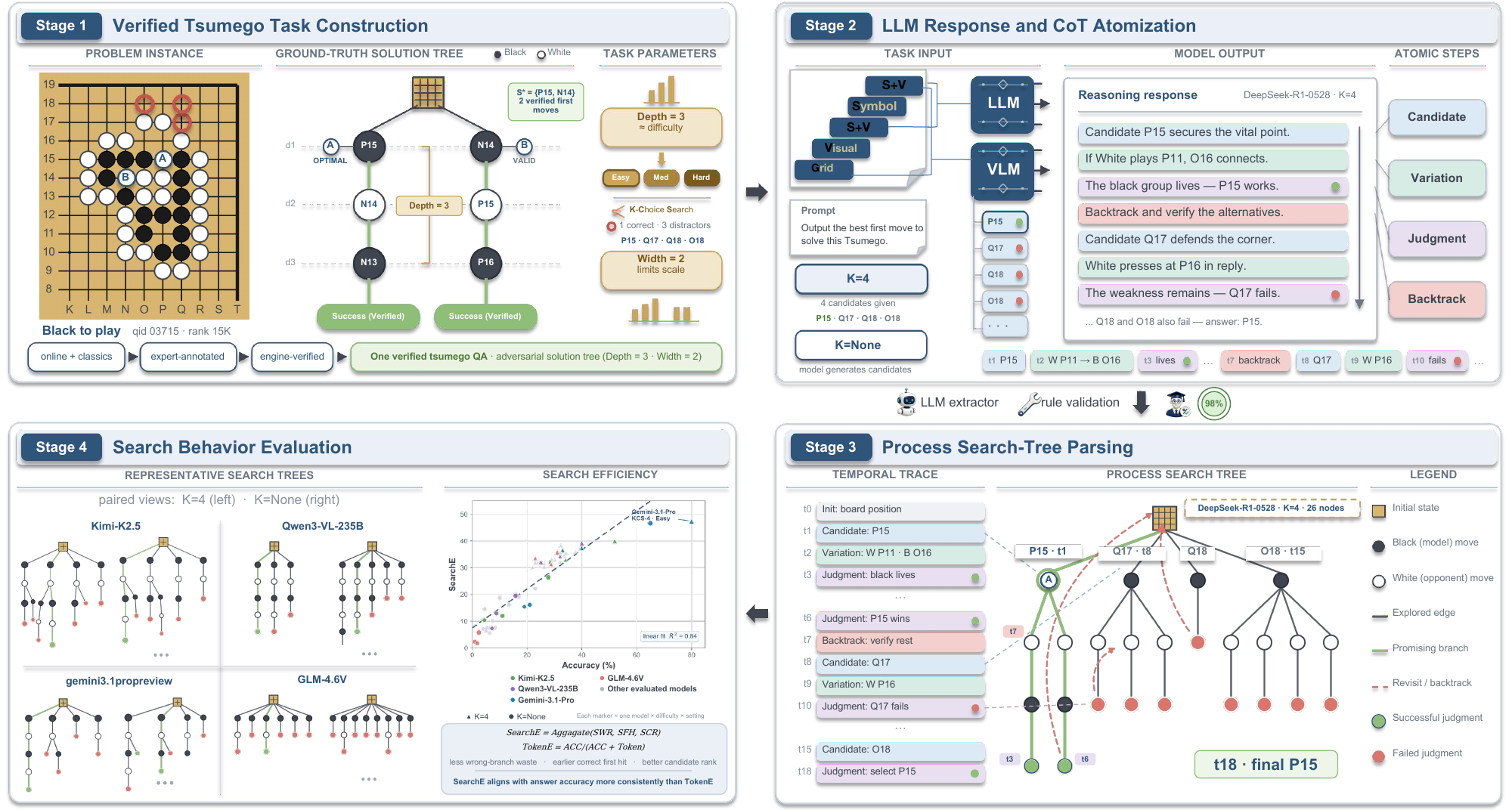}
  \vspace{-0.25em} 
  \caption{Overview of \tsumego{}: verified problems are rendered into five modalities, solved through bounded or open K-Search, parsed into process search trees, and evaluated by Search Efficiency with supporting diagnostics.}
  \label{fig:workflow}
\end{figure*}

Chain-of-Thought (CoT) and extended thinking have become central to LLM reasoning, improving performance on mathematics, code generation, and scientific problem solving~\cite{wei2022chain,openai2024o1,deepseek2025r1}.
This shift moves evaluation beyond final-answer correctness toward whether reasoning processes are effective and efficient.

Existing evaluations mainly study the quality and efficiency of reasoning along a single trajectory.
Traditional benchmarks such as GSM8K~\cite{cobbe2021gsm8k}, MATH~\cite{hendrycks2021math}, and ARC~\cite{clark2018arc} focus on final-answer correctness, while recent process-level methods analyze reasoning traces themselves.
CoTJudger measures necessary reasoning and structural redundancy from dependency graphs~\cite{li2026cotjudger}; ReEfBench maps traces into logical structures to analyze reasoning efficiency and behavioral patterns~\cite{fu2026reefbench}; and process-supervision or PRM-based methods evaluate intermediate steps through step-level feedback~\cite{lightman2024prm,song2025prmbench}.
These studies show whether a reasoning chain is correct, concise, and logically coherent, but mainly assume that effective reasoning follows a single derivation path, where the key question is which steps are necessary and which can be removed.

Many challenging problems require more than static problem solving along a single derivation path: models must explore alternatives, identify promising directions, and allocate resources across competing paths.
We call this capability \textbf{Search Efficiency} (\textbf{SearchE}): organizing reasoning search toward effective solutions across multiple possible trajectories.
Compared with Token Efficiency, which reflects observable token cost, SearchE is more internal: it tracks how models move through the solution space and redistribute effort across branches, rather than only how much surface text they produce.

Adversarial tasks naturally expose this ability because each decision must survive future responses, forcing comparison, verification, and revision.
Branching is therefore not inherently redundant; effective reasoning depends on whether exploration is organized around valuable paths.
Classical game AI and recent LLM search studies show the importance of explicit search~\cite{silver2016alphago,silver2017alphagozero,gandhi2024cognitive,feng2024alphazerolike}, but LLM benchmarks still lack a controlled way to evaluate search organization itself.

To address this gap, we introduce \tsumego{}, a benchmark based on Go life-and-death problems for evaluating SearchE in LLM reasoning.
The name draws on \emph{tsumego} (\emph{tsume-go}), the Japanese term for Go life-and-death problems, and emphasizes how models choose where to go next.
Tsumego offers a controlled adversarial environment with verifiable solution spaces, enabling analysis of both final answers and search over alternatives.
We parse free-form CoT into process search trees and report SearchE, which measures whether resources target the correct branch, and TokenE, which normalizes accuracy by observable token cost.
Experiments show that stronger models succeed by proposing the correct candidate earlier and sustaining effort on productive branches, establishing search organization as a dimension beyond outcome accuracy and token efficiency.

Our contributions are threefold:
\begin{itemize}[leftmargin=*,nosep]
  \item We introduce \tsumego{}, a process-level benchmark for evaluating Search Efficiency in controlled adversarial tasks with verifiable solution spaces~(\S\ref{sec:benchmark}).
  \item We propose a search-trace framework that converts free-form CoT into process search trees and diagnoses resource allocation with Search Efficiency and supporting trajectory/scale diagnostics~(\S\ref{sec:framework}).
  \item We show that Search Efficiency aligns more closely with search organization than token-level efficiency, and that \tsumego{} remains far from saturated: even frontier LLMs lag far behind neural-guided KataGo, making search organization a key underdeveloped direction for LLM reasoning~(\S\ref{sec:experiments}).
\end{itemize}

\section{Related Work}
\label{sec:related}

\paragraph{Reasoning Evaluation Benchmarks.}
Early reasoning benchmarks mainly evaluate final-answer accuracy.
GSM8K~\cite{cobbe2021gsm8k}, MATH~\cite{hendrycks2021math}, ARC~\cite{clark2018arc}, and BIG-Bench~\cite{srivastava2023bigbench} measure whether models can solve mathematical, scientific, and general reasoning problems.
Self-consistency further improves performance by aggregating multiple sampled solutions~\cite{wang2023selfconsistency}.
Formal and synthetic logic benchmarks further test rule-based inference and constraint satisfaction~\cite{han2024folio,saparov2023language,lin2025zebralogic,liu2025loginumsynth}. However, outcome-based evaluation cannot reveal how models reason.

Recent studies have begun to analyze intermediate reasoning traces. ROSCOE and ReCEval evaluate CoT quality~\cite{golovneva2023roscoe,prasad2023receval}; CoTJudger measures reasoning redundancy through dependency graphs and shortest effective paths~\cite{li2026cotjudger}; ReEfBench analyzes reasoning structures and efficiency~\cite{fu2026reefbench}; ProcessBench, PRMBench, and process reward modeling evaluate step-level correctness and supervision signals~\cite{zheng2025processbench,song2025prmbench,lightman2024prm}.
Other works study the cost, scaling behavior, inference-time behavior, and faithfulness of long CoT~\cite{li2025thinkbench,luo2025o1pruner,sui2025stop,chen2025overthinking,shen2025faithcot,parashar2025inference,gandhi2025cognitivebehaviors}.
These methods mainly evaluate the quality of a single reasoning trajectory. In contrast, \tsumego{} studies how models organize search across multiple possible trajectories.

\paragraph{Search-based Reasoning.}

Meanwhile, recent work improves LLM reasoning by introducing explicit search procedures. Tree-of-Thought and Graph-of-Thoughts represent intermediate thoughts as trees or graphs~\cite{yao2023treethought,besta2024graph}; RAP formulates reasoning as planning~\cite{hao2023reasoning}; LATS, AlphaZero-like search, Stream of Search, and LE-MCTS explore search-based inference and training~\cite{zhou2024language,feng2024alphazerolike,gandhi2024cognitive,park2025lemcts}.
These methods show that structured search can improve reasoning when supplied as an external procedure, but leave open whether LLMs can organize search within their own reasoning traces.
\tsumego{} evaluates this internal search-organization ability directly.

\paragraph{Adversarial Reasoning Environments.}
Adversarial environments provide structured spaces for studying planning and search.
AlphaGo and AlphaGo Zero demonstrate the importance of tree search for decision making in games~\cite{silver2016alphago,silver2017alphagozero}.
Recent LLM studies on chess, Othello, and Go investigate state tracking, move prediction, and strategic reasoning~\cite{toshniwal2022chess,li2023othello,ma2025mixing}.
These works mainly evaluate task performance or state understanding.
\tsumego{} uses Go life-and-death problems differently: as a controlled environment for analyzing how models explore, verify, and organize reasoning search.

\section{TsuGO}
\label{sec:benchmark}

\begin{figure*}[t]
  \centering
  \includegraphics[width=\textwidth]{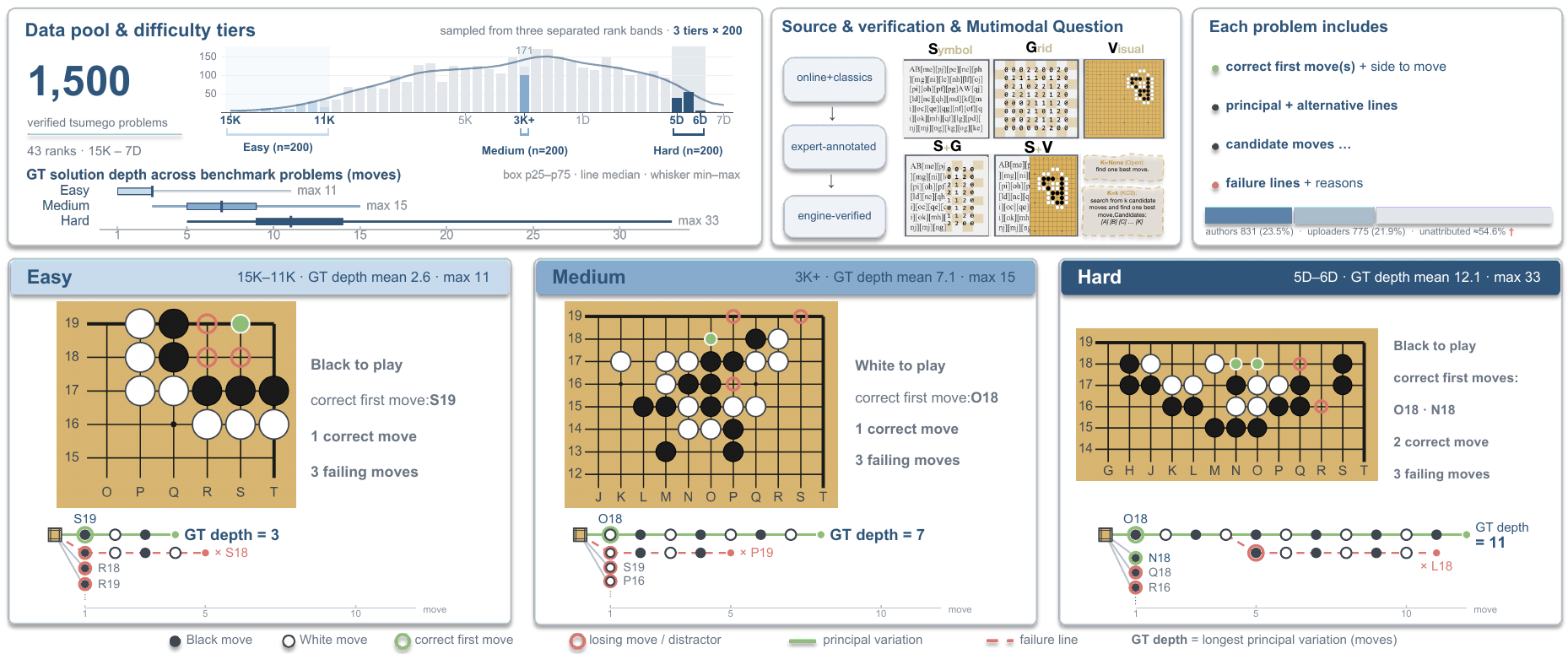}
  \vspace{-0.25em} 
  \caption{Overview of the \tsumego{} dataset construction and evaluation settings.}
  \label{fig:dataset_overview}
\end{figure*}

\subsection{Why Tsumego?}
\label{sec:why}
As introduced above, \tsumego{} uses Go life-and-death problems to study how models organize search rather than whether they know a final answer.
This setting is especially suitable because adversarial solution spaces are large and dynamic: each candidate move changes the space of possible replies, and a line remains valid only if it survives the opponent's strongest response.
Tsumego provides this structure in a compact and verifiable form: relevant moves are local, variations can be checked against reference solutions and engine analysis, and positions can be rendered as coordinates, matrices, or images.
We therefore use tsumego not to test Go strength, but as a closed adversarial environment where the difficulty of Go-style search comes from dynamically changing branches and the need to separate domain knowledge from search organization.

Table~\ref{tab:bench_compare} shows that prior benchmarks cover answer accuracy, step errors, or CoT redundancy, but not resource allocation across adversarial branches.
\tsumego{} uniquely combines process traces, reasoning topology, verifiable states, controlled candidates, multimodal inputs, branch search, and search-organization analysis.
\begin{table}[t]
\caption{Comparison with representative reasoning benchmarks and process-evaluation frameworks.
$\checkmark$ denotes a primary target, $\circ$ partial support, and $\times$ no explicit support.
T = trace, P = topology, V = verification, C = controlled candidate space, M = multimodal input, B = branch search, O = search organization.}
\label{tab:bench_compare}
\vspace{-0.20em} 
\centering
\footnotesize
\setlength{\tabcolsep}{2.2pt}
\renewcommand{\arraystretch}{1.03}
\begin{tabular*}{\columnwidth}
{@{\extracolsep{\fill}}
>{\raggedright\arraybackslash}p{0.46\columnwidth}
*{7}{>{\centering\arraybackslash}p{0.047\columnwidth}}
}
\toprule
\textbf{Benchmark group and examples} &
\textbf{T} &
\textbf{P} &
\textbf{V} &
\textbf{C} &
\textbf{M} &
\textbf{B} &
\textbf{O} \\
\midrule
\textit{Outcome} (GSM8K, MATH, ARC, BIG-Bench) &
$\circ$ & $\times$ & $\circ$ & $\circ$ & $\times$ & $\times$ & $\times$ \\
\textit{Logic} (FOLIO, PrOntoQA, ZebraLogic) &
$\circ$ & $\circ$ & \checkmark & \checkmark & $\times$ & $\circ$ & $\times$ \\
\textit{Process} (ROSCOE, ReCEval, ProcessBench, PRMB) &
\checkmark & $\circ$ & \checkmark & $\circ$ & $\times$ & $\times$ & $\times$ \\
\textit{CoT efficiency} (CoTJudger, ReEfBench) &
\checkmark & \checkmark & $\circ$ & $\circ$ & $\times$ & $\circ$ & $\circ$ \\
\textit{Behavior} (Sys2Bench, CogBehaviors) &
\checkmark & $\circ$ & $\circ$ & $\circ$ & $\times$ & $\circ$ & $\times$ \\
\textit{Board games} (Chess, Othello, Go) &
\checkmark & \checkmark & \checkmark & $\circ$ & $\times$ & $\circ$ & $\times$ \\
\midrule
\textbf{\tsumego{}} &
\textbf{\checkmark} & \textbf{\checkmark} & \textbf{\checkmark} &
\textbf{\checkmark} & \textbf{\checkmark} & \textbf{\checkmark} &
\textbf{\checkmark} \\
\bottomrule
\end{tabular*}
\end{table}

\subsection{Data Curation}
\label{sec:data}

Problems are curated from public tsumego materials, classical collections, and traceable source records, including texts such as \textit{Xuanxuan Qijing} and \textit{Go Life-and-Death Dictionary}.
To avoid reproducing platform-specific material, \tsumego{} keeps only normalized board states, candidate points, reference solution trees, and provenance records; each problem is stored in SGF and converted to JSON with a $19{\times}19$ matrix, coordinates, side to play, and solution tree.

We retain problems satisfying three criteria: \textbf{Locality}, where relevant stones occupy a bounded region; \textbf{First-move determinacy}, where correct first moves are unique or finite and not ko-dependent; and \textbf{Verifiability}, where ground-truth solutions are expert-checked and cross-validated by a Go engine.
Each solution tree stores the correct first move, principal variation, valid alternatives, refutations, and failure annotations for both answer scoring and search-trace evaluation~(\S\ref{sec:framework}); additional curation and traceability details appear in the appendix.

Each position is rendered as Symbolic coordinates, Grid matrices, Symbolic+Grid, Visual images, and Symbolic+Visual, enabling text-only and vision-language models to face structurally identical positions.
In bounded-candidate tests, distractors are locally salient but tactically wrong points, common amateur misconceptions, or apparent sente moves that allow refutation.

\subsection{Dataset Overview}
\label{sec:dataset_overview}

\tsumego{} is organized by difficulty, modality, and candidate-space setting.
The curated pool contains 1,500 problems in five input forms; the main evaluation uses 600 problems, sampled as 200 problems from each of three evaluated difficulty tiers, while harder remaining subsets are reserved for future models as capabilities improve.
Difficulty tiers combine human-rank bands, main-line depth, and plausible wrong candidates; Table~\ref{tab:difficulty} shows a progression from shallow kyu-level to dan-level amateur problems, with rank bands used only as coarse descriptors.

\begin{table}[t]
  \caption{Difficulty tiers in \tsumego{}. Human-rank bands are approximate descriptors; depth is measured on reference solution trees.}
  \label{tab:difficulty}
  \vspace{-0.20em} 
  \centering
  \small
  \setlength{\tabcolsep}{4pt}
  \renewcommand{\arraystretch}{1.03}
  \begin{tabular*}{\columnwidth}{@{\extracolsep{\fill}}llcc}
    \toprule
    Tier & Human rank band & Avg.\ depth & Max depth \\
    \midrule
    Easy         & 15K--11K & 2.8 &  7 \\
    Intermediate & 7K--6K   & 5.7 & 11 \\
    Medium       & 3K+      & 6.6 & 15 \\
    Hard         & 5D--6D   & 11.9 & 37 \\
    \bottomrule
  \end{tabular*}
\end{table}

We evaluate each problem under two candidate-space conditions: $K{=}4$, where the model selects from four first moves including the correct one, and $K{=}\mathrm{None}$, where it must generate, compare, and justify a move from the board.
This separates candidate discrimination from independent search organization across symbolic, grid, and visual modalities.
We also test rotation, reflection, color inversion, coordinate relabeling, and candidate-order permutation as robustness controls against surface memorization, coordinate shortcuts, and answer-order bias; details appear in the appendix.

\section{Search-Trace Analysis Framework}
\label{sec:framework}

\subsection{Trace Source}
\label{sec:trace_collection}
\tsumego{} uses each model's observable reasoning output to reconstruct search organization, using full CoT or reasoning content when available.
For closed-source models that expose only compressed summaries, we analyze those summaries as observable artifacts rather than complete internal computation.
They still reveal candidate proposal, variation reading, position judgment, and branch switching, although summary-length scale metrics are not comparable to full-CoT models.
Prompts provide only the board, side to play, and task instruction, without imposing a reasoning format or tree-search strategy.

\subsection{Process Search Tree}
\label{sec:pstree}

To compare free-form traces, \tsumego{} parses each output into a process search tree $T=(V,E,\tau)$, where $V$ is the node set, $E$ the directed edges, and $\tau$ a strictly increasing timestamp.
The root $r$ is the initial board state; action nodes $a$ are proposed or simulated moves with $\mathrm{side}(a)$ and $\mathrm{pos}(a)$; evaluation nodes $j$ are terminal judgments with $\mathrm{polarity}(j)\in\{\textit{win},\textit{lose},\textit{undetermined}\}$.

The root's action children form the first-level candidates $C=\{c_1,\ldots,c_m\}$.
For candidate $c$, $|\mathrm{subtree}(c)|$ measures allocated resources and maximum root-to-leaf distance measures reading depth.
Cross-candidate timestamp transitions are search jumps; jumps to previously visited candidates are backtracking.
Duplicate actions under the same parent are merged, while revisits under different branches remain separate to preserve search history.

\subsection{Extraction Pipeline}
\label{sec:parsing}

Algorithm~\ref{alg:extraction} summarizes extraction.
We build a domain prior dictionary from open-ended answers with four step types: candidate exploration, variation reading, position evaluation, and backtracking.
An LLM-based extractor then identifies action and evaluation nodes, classifies edges from temporal and semantic context, and closes branches with win/lose/undetermined judgments.
Structural validation enforces well-formed trees: roots connect only to actions, evaluations are leaves, unclosed branches receive undetermined leaves, repeated sibling actions are merged, and timestamps remain monotonic.
Final trees are selected from parallel extractors by rule filtering and cross-validation.
We further validate extraction on 300 sampled problems with two human experts assisted by KataGo; Gemini-2.5-Pro reaches 93--98\% agreement, with the range reflecting semantic ambiguity in free-form traces, and details appear in the appendix.

\begin{algorithm}[t]
\caption{Process Search Tree Extraction}\label{alg:extraction}
\begin{algorithmic}[1]
\REQUIRE Free-form reasoning text $\mathcal{T}$, prior dictionary $\mathcal{D}$
\ENSURE Process search tree $T = (V, E, \tau)$
\STATE Initialize $V \leftarrow \{r\}$, $E \leftarrow \emptyset$, $\tau(r)\leftarrow 0$
\STATE Scan $\mathcal{T}$ using $\mathcal{D}$ to identify atomic steps: \textsc{Explore}, \textsc{Read}, \textsc{Evaluate}, \textsc{Backtrack}
\FOR{each identified step in temporal order}
  \IF{step is \textsc{Explore} or \textsc{Read}}
    \STATE Create action node $a$; infer parent by edge classification; assign $\tau(a)$
  \ELSIF{step is \textsc{Evaluate}}
    \STATE Create evaluation node $j$; assign polarity and $\tau(j)$
  \ELSIF{step is \textsc{Backtrack}}
    \STATE Record branch switch and update current context
  \ENDIF
\ENDFOR
\STATE Apply structural validation and supplement undetermined leaves
\RETURN $T$
\end{algorithmic}
\vspace{-0.10em}
\end{algorithm}
\vspace{-0.40em}

\subsection{Metrics}
\label{sec:metrics}

\tsumego{} reports answer accuracy as first-move hit rate: a response is correct when its selected key move, i.e., the tsumego vital point, matches the reference first move.
Because accuracy cannot show how the model reaches, misses, or abandons that move, we define process metrics, including SearchE, on extracted trees to measure resource allocation across candidates.

To contrast token-centered process efficiency with search-organization efficiency, we highlight two signals:
\emph{SearchE} aggregates wrong-branch waste ($\swr$), first-hit behavior ($\sfh$), and correct-candidate search rank ($\scr$):
$SearchE=100\cdot(0.5(1-\swr)+0.3\,\sfh+0.2(1-\scr))$.
\emph{TokenE} measures accuracy relative to observable thinking-token cost: $TokenE=100\cdot A/(A+\itt/1000)$, with $A=100\cdot Acc$.
Full formulas and weight sensitivity appear in the appendix; we assign 0.3 to $\sfh$ and 0.2 to $\scr$ because both measure early correct-branch exploration, while $\sfh$ more directly reflects early correct-candidate intuition.

Due to space, the main text uses SearchE as the overall search-efficiency summary and reports the remaining metrics as diagnostics: (i) search metrics, including Search Waste Ratio ($\swr$), Search First Hit ($\sfh$), and Search Correct Rank ($\scr$); (ii) trajectory metrics, including Trajectory Max Depth ($\tmd$), Trajectory Max Fan-out ($\tmf$), and Trajectory Node Count ($\tnc$); and (iii) scale metrics, including Inference Total Tokens ($\itt$) and Inference Per Node ($\ipn$).
For proprietary summary-only models, scale metrics are summary-derived references rather than internal-compute measures.
Non-LLM baselines provide only comparable search-side metrics because they are not token-driven, and neural-guided KataGo does not expose internal decisions in a CoT-like form.
Full definitions, LLM parameters, and non-LLM reference settings appear in the appendix.

\section{Experiments}
\label{sec:experiments}

  \begin{table*}[!t]
    \centering
    \small
    \setlength{\tabcolsep}{4.7pt}
    \renewcommand{\arraystretch}{0.90}
    \setlength{\aboverulesep}{0.25ex}
    \setlength{\belowrulesep}{0.25ex}
    \setlength{\cmidrulesep}{0.12ex}
      \begin{tabular}{ll ccc|ccc|cc}
      \toprule
      \textbf{Model} & \textbf{Diff.} & \textbf{Acc} & \textbf{SearchE} & \textbf{TokenE} & \multicolumn{3}{c|}{\textbf{Trajectory}} & \multicolumn{2}{c}{\textbf{Scale}} \\
      \cmidrule(lr){4-4} \cmidrule(lr){5-5} \cmidrule(lr){6-8} \cmidrule(lr){9-10}
      & & & $\mathit{SearchE}\uparrow$ & $\mathit{TokenE}\uparrow$ & $\mathit{TMD}$ & $\mathit{TMF}$ & $\mathit{TNC}$ & $\mathit{IPN}\downarrow$ & $\mathit{ITT}(k)\downarrow$ \\
      \midrule
      \multicolumn{10}{l}{\textbf{\textit{Reasoning Models}}} \\
  \multirow{3}{*}{Kimi-K2.5} & Easy & \pairS{\eB{52.0}}{\eB{27.8}} & \pairS{\eB{39.9}}{\eB{26.4}} & $68.2\mkern5mu|\mkern5mu47.2$ & $4.6\mkern5mu|\mkern5mu4.8$ & $2.3\mkern5mu|\mkern5mu2.4$ & $20.5\mkern5mu|\mkern5mu24.7$
  & $1269\mkern5mu|\mkern5mu1176$ & $24.3\mkern5mu|\mkern5mu31.1$ \\
   & Med & \pairS{\mB{28.4}}{\mB{11.0}} & \pairS{\mB{32.3}}{12.0} & $56.0\mkern5mu|\mkern5mu27.7$ & $4.7\mkern5mu|\mkern5mu5.3$ & $2.1\mkern5mu|\mkern5mu2.5$ & $21.2\mkern5mu|\mkern5mu26.2$ & $1258\mkern5mu|\mkern5mu1014$
  & $22.3\mkern5mu|\mkern5mu28.7$ \\
   & Hard & \pairS{\hB{34.0}}{4.4} & \pairS{32.9}{10.4} & $61.5\mkern5mu|\mkern5mu12.2$ & $5.0\mkern5mu|\mkern5mu4.9$ & $2.3\mkern5mu|\mkern5mu2.2$ & $19.6\mkern5mu|\mkern5mu26.8$ & $1213\mkern5mu|\mkern5mu1197$ &
  $21.3\mkern5mu|\mkern5mu31.8$ \\
      \addlinespace[0pt]
  \multirow{3}{*}{\shortstack[l]{Qwen3-VL-235B-\\Thinking}} & Easy & \pairS{40.0}{15.8} & \pairS{39.1}{19.5} & $68.6\mkern5mu|\mkern5mu46.9$ & $5.1\mkern5mu|\mkern5mu5.1$ & $2.2\mkern5mu|\mkern5mu2.3$ &
  $24.4\mkern5mu|\mkern5mu27.3$ & $854\mkern5mu|\mkern5mu745$ & $18.3\mkern5mu|\mkern5mu17.9$ \\
   & Med & \pairS{25.0}{8.8} & \pairS{32.1}{12.9} & $58.4\mkern5mu|\mkern5mu32.2$ & $5.4\mkern5mu|\mkern5mu4.9$ & $2.0\mkern5mu|\mkern5mu2.1$ & $22.9\mkern5mu|\mkern5mu23.7$ & $815\mkern5mu|\mkern5mu858$ &
  $17.8\mkern5mu|\mkern5mu18.5$ \\
   & Hard & \pairS{32.0}{\hB{7.2}} & \pairS{34.2}{7.3} & $62.5\mkern5mu|\mkern5mu27.6$ & $5.4\mkern5mu|\mkern5mu5.0$ & $2.2\mkern5mu|\mkern5mu2.1$ & $23.2\mkern5mu|\mkern5mu25.0$ & $901\mkern5mu|\mkern5mu794$ &
  $19.2\mkern5mu|\mkern5mu18.9$ \\
      \addlinespace[0pt]
  \multirow{3}{*}{\shortstack[l]{Qwen3-VL-30B-\\Thinking}} & Easy & \pairS{34.2}{10.0} & \pairS{35.4}{18.8} & $65.6\mkern5mu|\mkern5mu34.1$ & $3.9\mkern5mu|\mkern5mu3.7$ & $1.5\mkern5mu|\mkern5mu1.4$ &
  $15.6\mkern5mu|\mkern5mu17.5$ & $1216\mkern5mu|\mkern5mu1264$ & $17.9\mkern5mu|\mkern5mu19.3$ \\
   & Med & \pairS{23.0}{4.6} & \pairS{32.0}{\mB{14.5}} & $56.1\mkern5mu|\mkern5mu19.3$ & $3.7\mkern5mu|\mkern5mu3.6$ & $1.4\mkern5mu|\mkern5mu1.2$ & $14.7\mkern5mu|\mkern5mu17.3$ & $1285\mkern5mu|\mkern5mu1304$ &
  $18.0\mkern5mu|\mkern5mu19.2$ \\
   & Hard & \pairS{30.2}{3.6} & \pairS{32.8}{\hB{10.5}} & $62.7\mkern5mu|\mkern5mu15.8$ & $3.8\mkern5mu|\mkern5mu3.4$ & $1.5\mkern5mu|\mkern5mu1.2$ & $15.2\mkern5mu|\mkern5mu17.7$ & $1259\mkern5mu|\mkern5mu1256$ &
  $18.0\mkern5mu|\mkern5mu19.2$ \\
      \addlinespace[0pt]
  \multirow{3}{*}{\shortstack[l]{MiniMax-M2.5\\{\scriptsize (Non-VL)}}} & Easy & \pairS{37.7}{13.3} & \pairS{36.0}{16.1} & $\eB{76.8}\mkern5mu|\mkern5mu\eB{54.5}$ & $3.5\mkern5mu|\mkern5mu3.5$ &
  $1.7\mkern5mu|\mkern5mu1.6$ & $15.0\mkern5mu|\mkern5mu18.3$ & $827\mkern5mu|\mkern5mu687$ & $\eB{11.4}\mkern5mu|\mkern5mu\eB{11.1}$ \\
   & Med & \pairS{22.0}{7.3} & \pairS{30.1}{10.0} & $\mB{80.9}\mkern5mu|\mkern5mu\mB{38.6}$ & $3.1\mkern5mu|\mkern5mu3.5$ & $1.2\mkern5mu|\mkern5mu1.3$ & $12.6\mkern5mu|\mkern5mu17.3$ & $\mB{540}\mkern5mu|\mkern5mu740$ &
  $\mB{5.2}\mkern5mu|\mkern5mu\mB{11.6}$ \\
   & Hard & \pairS{26.3}{6.3} & \pairS{31.5}{7.4} & $\hB{73.5}\mkern5mu|\mkern5mu\hB{33.5}$ & $3.3\mkern5mu|\mkern5mu3.3$ & $1.5\mkern5mu|\mkern5mu1.4$ & $14.5\mkern5mu|\mkern5mu12.9$ & $734\mkern5mu|\mkern5mu722$ &
  $\hB{9.5}\mkern5mu|\mkern5mu\hB{12.5}$ \\
      \addlinespace[0pt]
  \multirow{3}{*}{\shortstack[l]{DeepSeek-R1-\\0528{\scriptsize (Non-VL)}}} & Easy & \pairS{32.3}{17.0} & \pairS{38.3}{19.1} & $71.1\mkern5mu|\mkern5mu53.8$ & $4.6\mkern5mu|\mkern5mu4.5$ & $2.0\mkern5mu|\mkern5mu2.1$ &
  $19.5\mkern5mu|\mkern5mu23.3$ & $\eB{743}\mkern5mu|\mkern5mu\eB{676}$ & $13.1\mkern5mu|\mkern5mu14.6$ \\
   & Med & \pairS{26.3}{8.3} & \pairS{30.2}{7.6} & $70.3\mkern5mu|\mkern5mu37.4$ & $4.6\mkern5mu|\mkern5mu4.3$ & $2.1\mkern5mu|\mkern5mu2.1$ & $21.2\mkern5mu|\mkern5mu22.6$ & $547\mkern5mu|\mkern5mu\mB{680}$ &
  $11.1\mkern5mu|\mkern5mu13.9$ \\
   & Hard & \pairS{33.7}{5.3} & \pairS{\hB{34.3}}{7.6} & $72.2\mkern5mu|\mkern5mu27.5$ & $4.6\mkern5mu|\mkern5mu4.4$ & $2.2\mkern5mu|\mkern5mu2.0$ & $21.6\mkern5mu|\mkern5mu21.1$ & $\hB{629}\mkern5mu|\mkern5mu\hB{691}$ &
  $13.0\mkern5mu|\mkern5mu14.0$ \\
      \midrule
      \multicolumn{10}{l}{\textbf{\textit{Non-reasoning Models}}} \\
  \multirow{3}{*}{GLM-4.6V} & Easy & \pairS{31.0}{2.4} & \pairS{35.8}{5.8} & $88.3\mkern5mu|\mkern5mu36.9$ & $2.7\mkern5mu|\mkern5mu2.5$ & $0.7\mkern5mu|\mkern5mu0.6$ & $11.5\mkern5mu|\mkern5mu9.9$ &
  $447\mkern5mu|\mkern5mu715$ & $4.1\mkern5mu|\mkern5mu4.1$ \\
   & Med & \pairS{23.0}{1.0} & \pairS{\mB{33.5}}{2.3} & $89.5\mkern5mu|\mkern5mu14.3$ & $2.7\mkern5mu|\mkern5mu2.5$ & $0.7\mkern5mu|\mkern5mu0.4$ & $11.4\mkern5mu|\mkern5mu8.6$ & $274\mkern5mu|\mkern5mu901$ &
  $2.7\mkern5mu|\mkern5mu6.0$ \\
   & Hard & \pairS{28.8}{1.8} & \pairS{31.3}{1.7} & $91.7\mkern5mu|\mkern5mu34.0$ & $2.7\mkern5mu|\mkern5mu2.5$ & $0.6\mkern5mu|\mkern5mu0.5$ & $10.9\mkern5mu|\mkern5mu9.7$ & $289\mkern5mu|\mkern5mu582$ &
  $2.6\mkern5mu|\mkern5mu3.5$ \\
      \addlinespace[0pt]
  \multirow{3}{*}{DeepSeek-V3.2} & Easy & \pairS{\eB{39.3}}{\eB{16.0}} & \pairS{\eB{37.0}}{\eB{20.0}} & $\eB{94.2}\mkern5mu|\mkern5mu\eB{85.1}$ & $4.7\mkern5mu|\mkern5mu5.0$ & $1.8\mkern5mu|\mkern5mu1.8$ &
  $17.0\mkern5mu|\mkern5mu16.4$ & $\eB{162}\mkern5mu|\mkern5mu\eB{220}$ & $\eB{2.4}\mkern5mu|\mkern5mu\eB{2.8}$ \\
   & Med & \pairS{\mB{24.0}}{\mB{7.0}} & \pairS{31.8}{\mB{7.4}} & $\mB{91.6}\mkern5mu|\mkern5mu\mB{72.2}$ & $4.6\mkern5mu|\mkern5mu4.9$ & $1.6\mkern5mu|\mkern5mu1.9$ & $16.3\mkern5mu|\mkern5mu16.0$ &
  $\mB{158}\mkern5mu|\mkern5mu\mB{225}$ & $\mB{2.2}\mkern5mu|\mkern5mu\mB{2.7}$ \\
   & Hard & \pairS{\hB{32.0}}{\hB{6.3}} & \pairS{\hB{32.2}}{\hB{5.5}} & $\hB{94.1}\mkern5mu|\mkern5mu\hB{70.8}$ & $4.6\mkern5mu|\mkern5mu4.9$ & $1.7\mkern5mu|\mkern5mu1.9$ & $17.2\mkern5mu|\mkern5mu16.2$ &
  $\stmax{\hB{138}}\mkern5mu|\mkern5mu\stmax{\hB{206}}$ & $\stmax{\hB{2.0}}\mkern5mu|\mkern5mu\stmax{\hB{2.6}}$ \\
      \midrule
      \multicolumn{10}{l}{\textbf{\textit{Proprietary Models}}} \\
  \multirow{3}{*}{Gemini-2.5-Flash} & Easy & \pairS{32.0}{23.0} & \pairS{35.7}{22.2} & $95.5\mkern5mu|\mkern5mu93.1$ & $3.4\mkern5mu|\mkern5mu4.1$ & $1.2\mkern5mu|\mkern5mu1.8$ & $13.8\mkern5mu|\mkern5mu16.0$ &
  \textcolor{gray}{$109\mkern5mu|\mkern5mu136$} & \textcolor{gray}{$1.5\mkern5mu|\mkern5mu1.7$} \\
   & Med & \pairS{29.0}{4.0} & \pairS{33.1}{7.2} & $94.2\mkern5mu|\mkern5mu70.2$ & $4.6\mkern5mu|\mkern5mu4.3$ & $2.1\mkern5mu|\mkern5mu1.8$ & $19.7\mkern5mu|\mkern5mu15.6$ & \textcolor{gray}{$102\mkern5mu|\mkern5mu146$} &
  \textcolor{gray}{$1.8\mkern5mu|\mkern5mu1.7$} \\
   & Hard & \pairS{29.0}{5.0} & \pairS{32.1}{6.4} & $94.8\mkern5mu|\mkern5mu74.6$ & $4.6\mkern5mu|\mkern5mu4.1$ & $2.1\mkern5mu|\mkern5mu2.1$ & $19.2\mkern5mu|\mkern5mu16.1$ & \textcolor{gray}{$98\mkern5mu|\mkern5mu133$} &
  \textcolor{gray}{$1.6\mkern5mu|\mkern5mu1.7$} \\
      \addlinespace[0pt]
      \multirow{3}{*}{\shortstack[l]{Gemini-3.1-Pro-\\Preview}} & Easy & \pairS{\eB{80.0}}{\eB{65.0}} & \pairS{\eB{47.3}}{\eB{46.6}} & $\eB{99.0}\mkern5mu|\mkern5mu\eB{98.9}$ & $4.1\mkern5mu|\mkern5mu3.9$ &
  $1.5\mkern5mu|\mkern5mu1.5$ & $13.7\mkern5mu|\mkern5mu9.6$ & \textcolor{gray}{$66\mkern5mu|\mkern5mu101$} & \textcolor{gray}{$0.8\mkern5mu|\mkern5mu0.7$} \\
   & Med & \pairS{\mB{40.0}}{\mB{21.0}} & \pairS{\mB{37.3}}{\mB{16.1}} & $\mB{98.5}\mkern5mu|\mkern5mu\mB{95.9}$ & $4.2\mkern5mu|\mkern5mu4.3$ & $1.7\mkern5mu|\mkern5mu1.6$ & $15.3\mkern5mu|\mkern5mu11.5$ &
  \textcolor{gray}{$52\mkern5mu|\mkern5mu104$} & \textcolor{gray}{$0.6\mkern5mu|\mkern5mu0.9$} \\
   & Hard & \pairS{33.0}{\hB{19.0}} & \pairS{\hB{36.5}}{\hB{15.4}} & $\stmax{\hB{97.9}}\mkern5mu|\mkern5mu\stmax{\hB{96.4}}$ & $4.0\mkern5mu|\mkern5mu4.3$ & $1.5\mkern5mu|\mkern5mu1.7$ & $14.3\mkern5mu|\mkern5mu11.1$ &
  \textcolor{gray}{$59\mkern5mu|\mkern5mu86$} & \textcolor{gray}{$0.7\mkern5mu|\mkern5mu0.7$} \\
      \midrule
      \multicolumn{7}{l}{\textbf{\textit{Search Baselines}}} & \multicolumn{1}{c}{\textit{TNC(k)}} & & \multicolumn{1}{c}{\textit{Budget}} \\
      \multirow{3}{*}{MCTS/UCT} & Easy & \pairS{33.0}{5.0} & \pairS{35.8}{13.1} & $\text{--}\mkern5mu|\mkern5mu\text{--}$ & $5.3\mkern5mu|\mkern5mu4.8$ & $45.9\mkern5mu|\mkern5mu47.1$ & $7.2\mkern5mu|\mkern5mu7.8$ &
  $\text{--}\mkern5mu|\mkern5mu\text{--}$ & $200^\dagger\mkern5mu|\mkern5mu200^\dagger$ \\
       & Med & \pairS{32.0}{1.0} & \pairS{35.1}{14.1} & $\text{--}\mkern5mu|\mkern5mu\text{--}$ & $5.2\mkern5mu|\mkern5mu4.8$ & $52.0\mkern5mu|\mkern5mu54.6$ & $8.7\mkern5mu|\mkern5mu9.1$ &
  $\text{--}\mkern5mu|\mkern5mu\text{--}$ & $200^\dagger\mkern5mu|\mkern5mu200^\dagger$ \\
       & Hard & \pairS{22.0}{1.0} & \pairS{30.8}{13.4} & $\text{--}\mkern5mu|\mkern5mu\text{--}$ & $5.3\mkern5mu|\mkern5mu4.9$ & $63.1\mkern5mu|\mkern5mu65.3$ & $11.0\mkern5mu|\mkern5mu11.4$ &
  $\text{--}\mkern5mu|\mkern5mu\text{--}$ & $200^\dagger\mkern5mu|\mkern5mu200^\dagger$ \\
      \addlinespace[0pt]
      \multirow{3}{*}{KataGo-b18} & Easy & \pairS{\eB{97.0}}{\eB{49.0}} & \pairS{\eB{96.9}}{\eB{53.3}} & $\text{--}\mkern5mu|\mkern5mu\text{--}$ & $\text{--}\mkern5mu|\mkern5mu\text{--}$ &
  $\text{--}\mkern5mu|\mkern5mu\text{--}$ & $\text{--}\mkern5mu|\mkern5mu\text{--}$ & $\text{--}\mkern5mu|\mkern5mu\text{--}$ & $200^\dagger\mkern5mu|\mkern5mu200^\dagger$ \\
       & Med & \pairS{\mB{75.0}}{\mB{46.0}} & \pairS{\mB{73.9}}{\mB{47.8}} & $\text{--}\mkern5mu|\mkern5mu\text{--}$ & $\text{--}\mkern5mu|\mkern5mu\text{--}$ & $\text{--}\mkern5mu|\mkern5mu\text{--}$ &
  $\text{--}\mkern5mu|\mkern5mu\text{--}$ & $\text{--}\mkern5mu|\mkern5mu\text{--}$ & $200^\dagger\mkern5mu|\mkern5mu200^\dagger$ \\
       & Hard & \pairS{\stmax{\hB{58.0}}}{\stmax{\hB{43.0}}} & \pairS{\stmax{\hB{57.0}}}{\stmax{\hB{41.5}}} & $\text{--}\mkern5mu|\mkern5mu\text{--}$ & $\text{--}\mkern5mu|\mkern5mu\text{--}$ & $\text{--}\mkern5mu|\mkern5mu\text{--}$ &
  $\text{--}\mkern5mu|\mkern5mu\text{--}$ & $\text{--}\mkern5mu|\mkern5mu\text{--}$ & $200^\dagger\mkern5mu|\mkern5mu200^\dagger$ \\
      \bottomrule
    \end{tabular}
    \vspace{-0.25em} 
    \caption{Compact main results. Cells report $K{=}4|K{=}\mathrm{None}$; Acc averages available modalities, and trace metrics use Symbolic input. SearchE aggregates SWR/SFH/SCR; TokenE is accuracy per token cost. $\itt$ is in thousand
  tokens; for search baselines, TNC is reported in thousands and the final scale column reports playout/visit budget. Best within each model-type group and each setting is marked per difficulty: \underline{\textbf{bold underline}}~=~Easy, \underline{underline}~=~Med, \textbf{bold}~=~Hard. $^\dagger$ denotes playout/visit budget; gray proprietary values are summary-derived
  and excluded from ranking. $^{*}$ marks the single best value in each column over the whole table, regardless of model type (shown for Hard).}
    \label{tab:main_results}
  \end{table*}

\subsection{Setup}
\label{sec:setup}

We evaluate open reasoning, open non-reasoning, and proprietary models on 600 sampled problems, with 200 per difficulty tier, under $K{=}4$ candidate selection and $K{=}\mathrm{None}$ generation.
The evaluated LLMs cover Moonshot AI Kimi-K2.5~\cite{kimi2026k25}, Alibaba Qwen3-VL~\cite{qwen2025vl}, MiniMax-M2.5~\cite{minimax2026m2}, DeepSeek-R1/V3.2~\cite{deepseek2025r1,deepseek2024v3}, Zhipu AI GLM-4.6V~\cite{zhipu2026glm46v}, and Google Gemini models~\cite{google2025gemini,google2026gemini31pro}.
Accuracy is averaged over available modalities, while process metrics use Symbolic input; for proprietary models, scale metrics are summary-derived references.

We use the Smargo implementation of MCTS/UCT~\cite{kocsis2006bandit,sun2022smargo} and KataGo~\cite{wu2019accelerating} as non-LLM references with 200 playouts or visits per problem; details appear in the appendix.
MCTS represents unguided search and KataGo neural-guided search; they locate LLM search organization but are not token-equivalent.

Beyond accuracy, the main table reports SearchE, our primary structural-efficiency signal, and TokenE, accuracy relative to thinking-token cost.

\subsection{Main Results}
\label{sec:main_results}

\textbf{Current LLMs are still far from stable \tsumego{} solving.}
Table~\ref{tab:main_results} shows accuracy dropping from constrained easy to open hard problems across systems.
Even under $K{=}4$ Easy, the strongest open model, Kimi-K2.5, reaches only 52.0 accuracy; Gemini-3.1-Pro reaches 80.0 but falls to 19.0 on $K{=}\mathrm{None}$ Hard.
Model classes show a clear but non-absolute ordering: Gemini models lead on easy settings but lose advantage with difficulty, while open reasoning models generally outperform weaker non-reasoning models in open search.
DeepSeek-V3.2, though not a long-reasoning model, approaches or exceeds some reasoning models in accuracy and SearchE; GLM-4.6V retains limited $K{=}4$ discrimination but nearly collapses under $K{=}\mathrm{None}$.
Size alone is insufficient: Qwen3-VL-235B usually beats its 30B variant, yet both degrade sharply on open hard problems, pointing to search organization rather than parameter count or token volume.

\textbf{The drop from $K{=}4$ to $K{=}\mathrm{None}$ exposes the open-search bottleneck.}
With $K{=}4$, the model mainly discriminates among provided options; with $K{=}\mathrm{None}$, it must generate, rank, and verify candidates itself.
The gap shows that many models use partial board-shape knowledge with candidates but struggle to preserve the correct direction in open space.
For example, Kimi-K2.5 drops from 52.0 accuracy and 39.9 SearchE on $K{=}4$ Easy to 27.8 and 26.4 on $K{=}\mathrm{None}$ Easy; GLM-4.6V nearly collapses across the open tiers, with SearchE approaching blind-search levels.

\textbf{Difficulty trends show that candidate constraints can mask open-search failures.}
Under $K{=}4$, candidate lists compress the search space, so some models do not degrade from Medium to Hard.
Once candidates are removed, difficulty produces a more consistent accuracy and SearchE decline, so the open setting better exposes independent search organization.
\begin{figure*}[!t]
  \centering
  \begin{subfigure}[t]{0.232\textwidth}
    \centering
    \includegraphics[width=\textwidth,height=0.155\textheight,keepaspectratio]{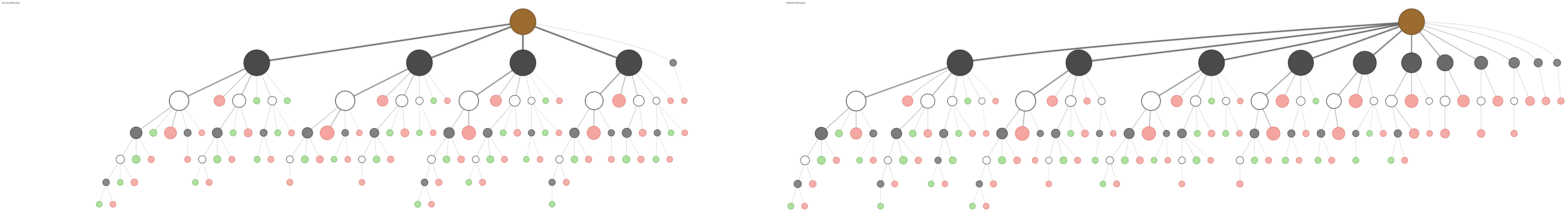}
    \caption{Kimi-K2.5}
  \end{subfigure}\hfill
  \begin{subfigure}[t]{0.232\textwidth}
    \centering
    \includegraphics[width=\textwidth,height=0.155\textheight,keepaspectratio]{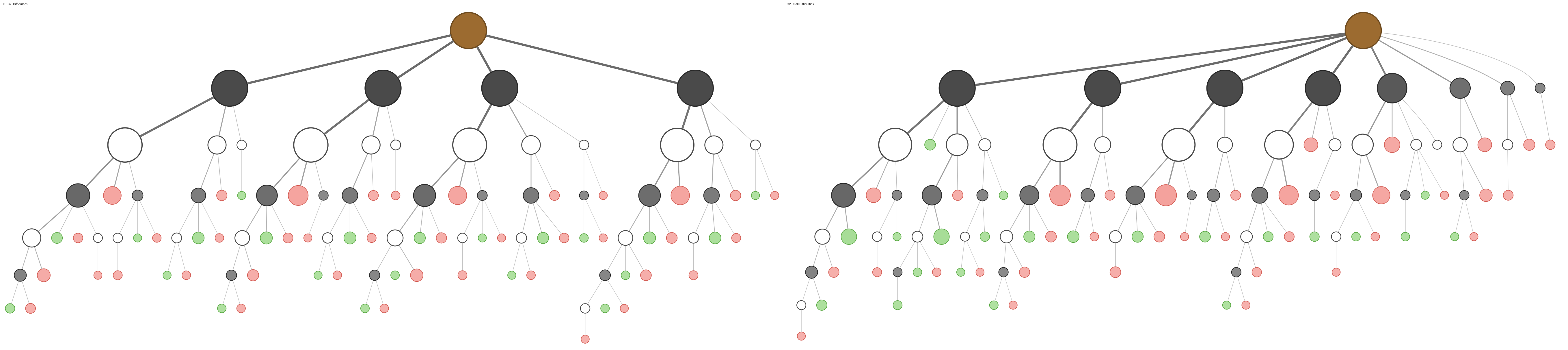}
    \caption{Qwen3-VL-235B-Thinking}
  \end{subfigure}\hfill
  \begin{subfigure}[t]{0.232\textwidth}
    \centering
    \includegraphics[width=\textwidth,height=0.155\textheight,keepaspectratio]{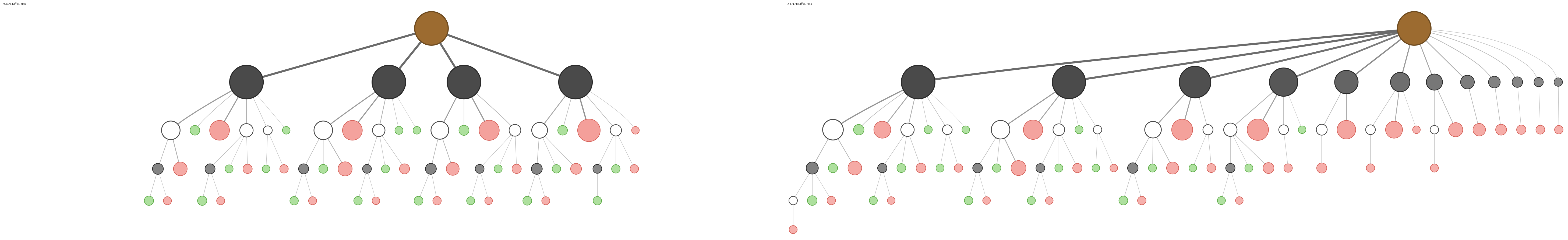}
    \caption{MiniMax-M2.5}
  \end{subfigure}\hfill
  \begin{subfigure}[t]{0.232\textwidth}
    \centering
    \includegraphics[width=\textwidth,height=0.155\textheight,keepaspectratio]{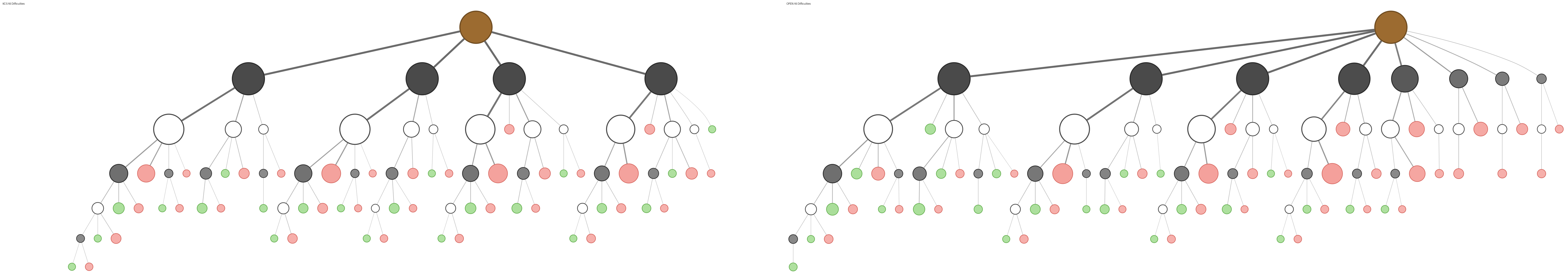}
    \caption{DeepSeek-R1-0528}
  \end{subfigure}\\[2pt]
  \begin{subfigure}[t]{0.232\textwidth}
    \centering
    \includegraphics[width=\textwidth,height=0.155\textheight,keepaspectratio]{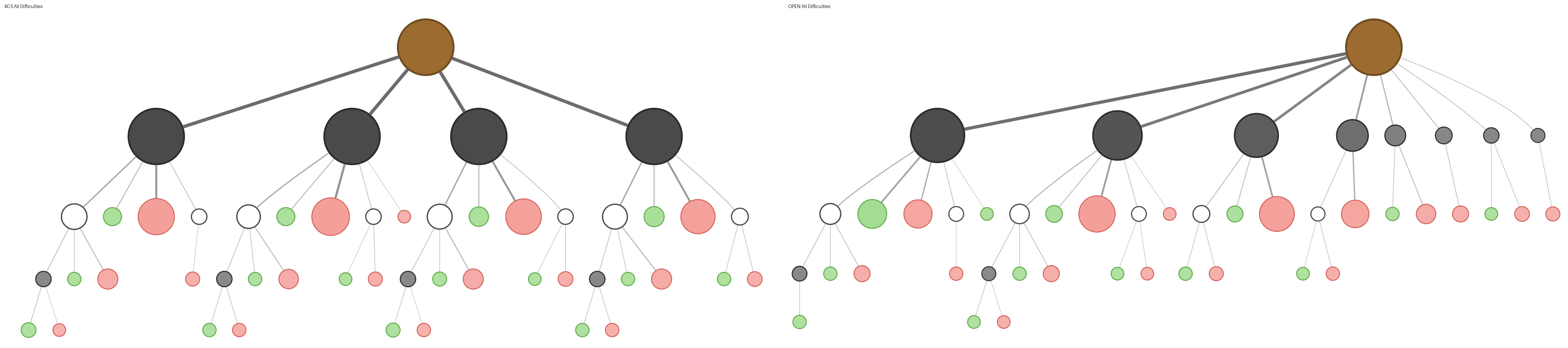}
    \caption{GLM-4.6V}
  \end{subfigure}\hfill
  \begin{subfigure}[t]{0.232\textwidth}
    \centering
    \includegraphics[width=\textwidth,height=0.155\textheight,keepaspectratio]{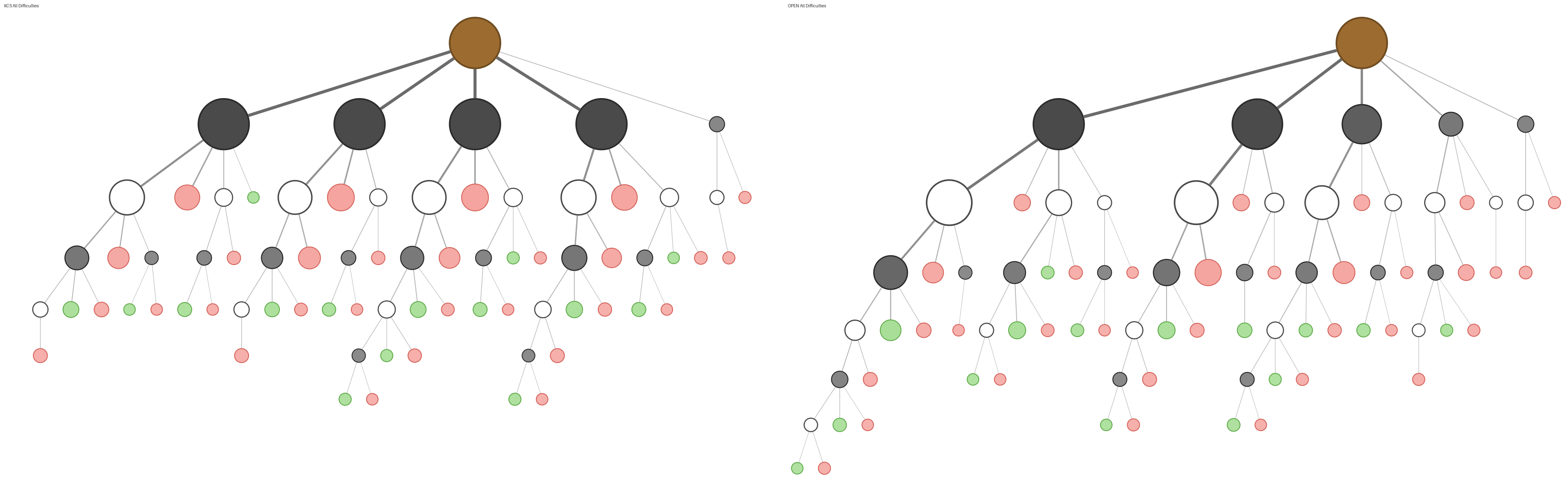}
    \caption{DeepSeek-V3.2$\dagger$}
  \end{subfigure}\hfill
  \begin{subfigure}[t]{0.232\textwidth}
    \centering
    \includegraphics[width=\textwidth,height=0.155\textheight,keepaspectratio]{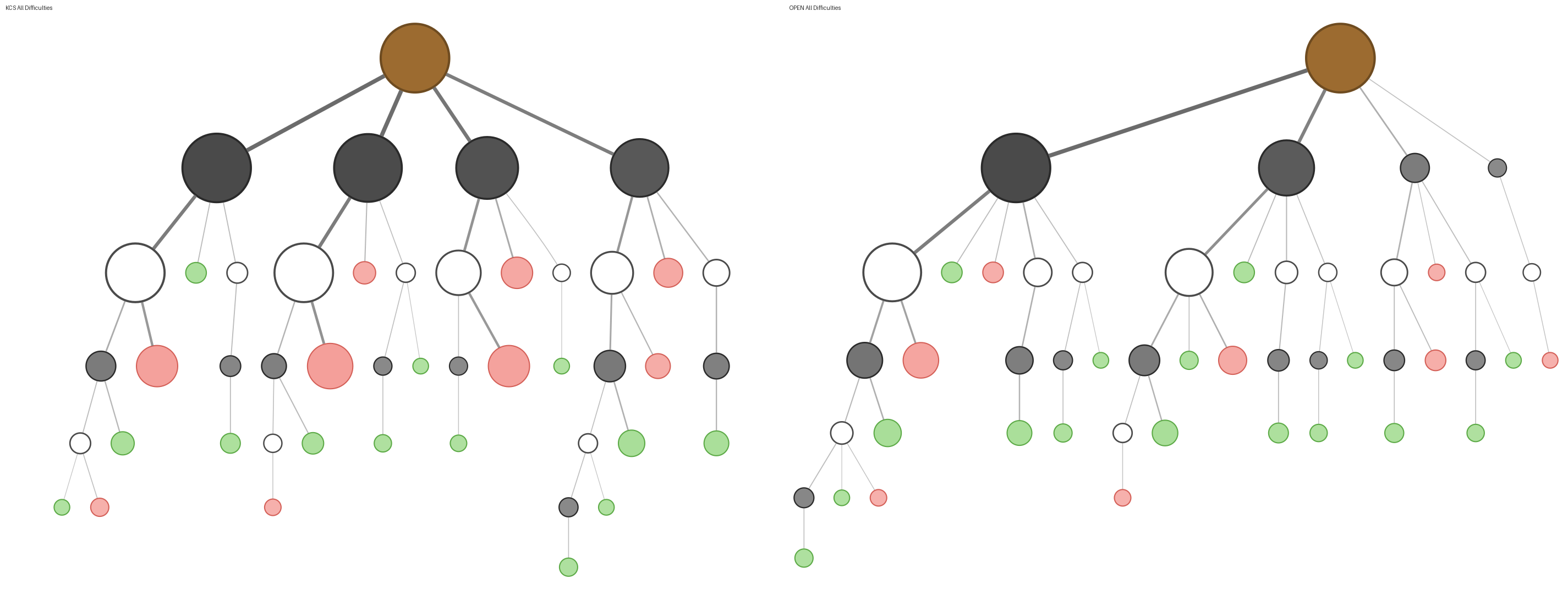}
    \caption{Gemini-3.1-Pro}
  \end{subfigure}\hfill
  \begin{subfigure}[t]{0.232\textwidth}
    \centering
    \includegraphics[width=\textwidth,height=0.155\textheight,keepaspectratio]{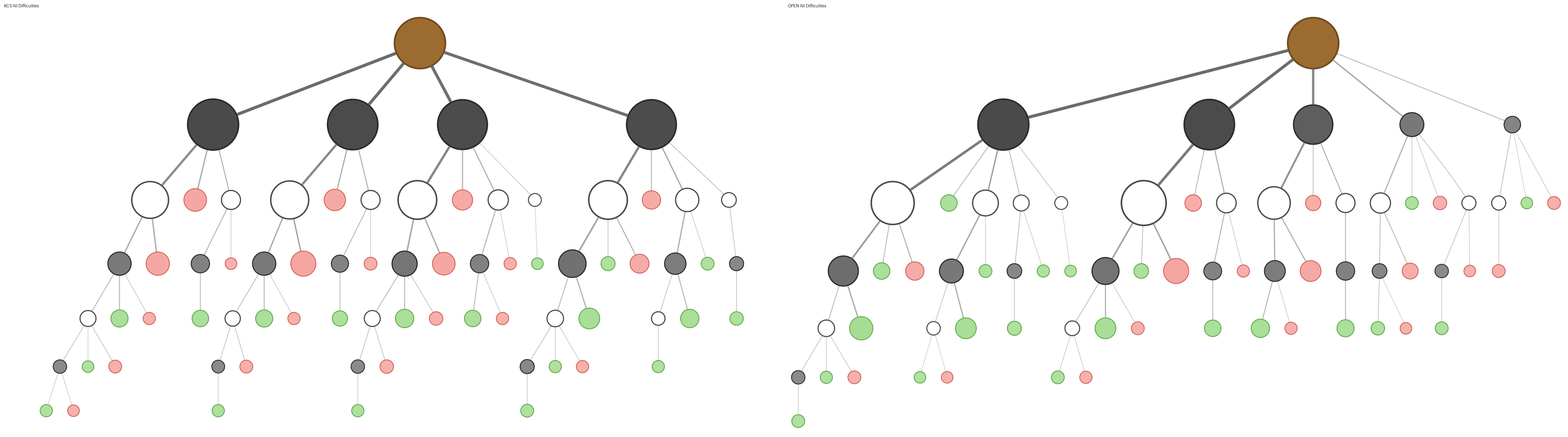}
    \caption{Gemini-2.5-Flash}
  \end{subfigure}
  \vspace{-0.30em} 
  \caption{Aggregated process search trees. Each panel shows $K{=}4$ (left) and $K{=}\mathrm{None}$ (right); node size is average subtree weight and color denotes polarity (green = win, red = lose).}
  \label{fig:treemap}
\end{figure*}

\textbf{Treemaps reveal focused and diffuse search.}
Figure~\ref{fig:treemap} visualizes aggregate process trees, where node size indicates branch resources.
Under $K{=}4$, candidate constraints make tree shapes similar.
Under $K{=}\mathrm{None}$, Kimi-K2.5 and Gemini-3.1-Pro concentrate on fewer candidates and push them toward verification, while GLM-4.6V and MCTS spread resources across shallow wrong branches.
Trajectory metrics in Table~\ref{tab:main_results} quantify these structures: $\tmd$ measures depth, $\tmf$ single-node fan-out, and $\tnc$ node count.
This matches SearchE: tree width or depth matters less than whether resources point toward the correct direction.

\subsection{Insight}
\label{sec:insight}

\textbf{Failure is a failure of search-resource allocation.}
The $K{=}4$ results show that many models retain board-shape knowledge when candidates are provided.
In the open setting, however, the correct move is often absent or displaced by shallow checks of other branches.
Failure is therefore not only missing the right move; often, the right candidate lacks early and sustained resources.
Tsumego requires adversarial verification around key candidates, not frequent switching among superficially plausible branches.

\begin{figure}[t]
  \vspace{-0.50em}
  \centering
  \includegraphics[width=\columnwidth]{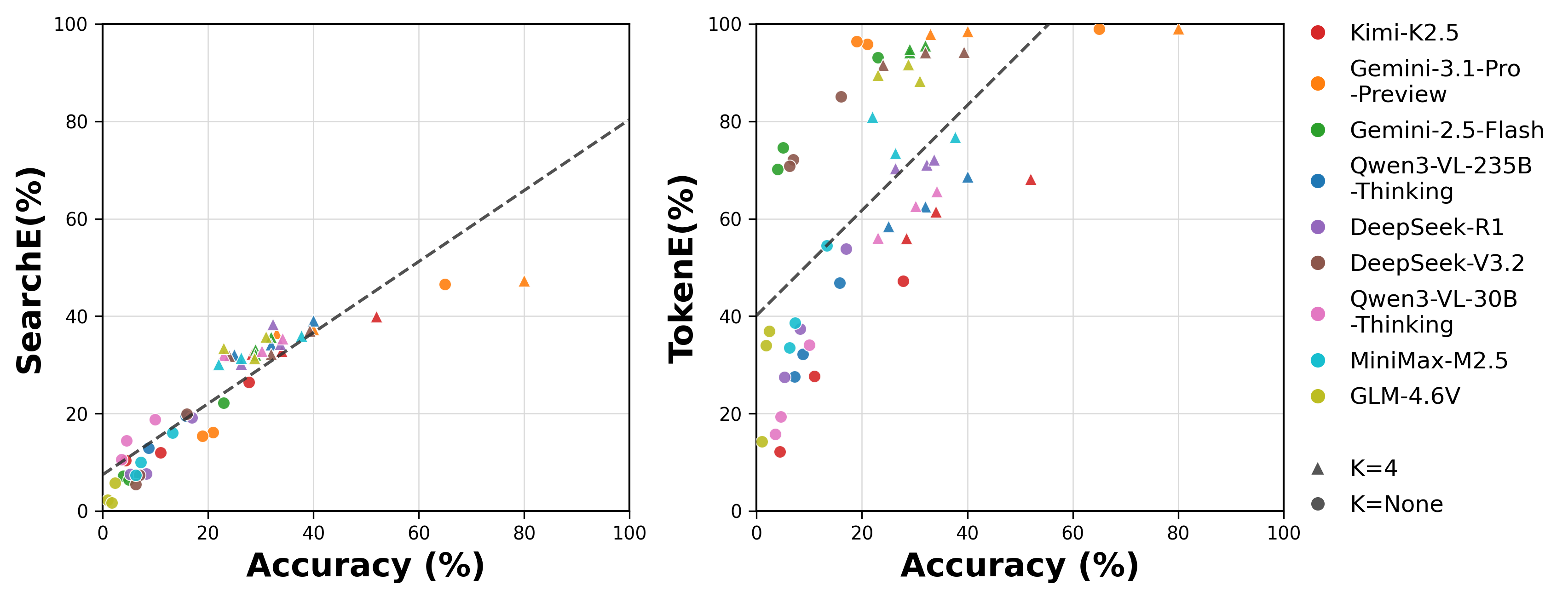}
  \vspace{-0.30em} 
  \caption{SearchE aligns with accuracy more consistently than TokenE.}
  \label{fig:correlation}
  \vspace{-0.50em}
\end{figure}

\textbf{SearchE is a stronger process signal than TokenE.}
Figure~\ref{fig:correlation} shows tighter clustering around accuracy for SearchE than TokenE, indicating stronger alignment with task success.
SearchE tracks accuracy more closely because stronger settings allocate more resources to the correct candidate.
TokenE is more dispersed, since low token cost does not imply effective search; SearchE better captures search intelligence, while TokenE mainly reflects cost.

\textbf{LLMs lie between blind and guided search.}
MCTS reaches mid-tier LLM accuracy under $K{=}4$ but drops to single digits without candidates, showing that unguided expansion cannot maintain direction in open space.
KataGo remains much stronger under the same visit budget, confirming that the task is solvable by well-guided search.
Strong LLMs outperform blind search, suggesting useful priors, but their gap to KataGo shows these priors are not yet stable, problem-adaptive search control.

\subsection{Further Analysis}
\label{sec:further_analysis}

\begin{figure}[t]
  \vspace{-0.50em}
  \centering
  \includegraphics[width=\columnwidth,height=0.22\textheight,keepaspectratio]{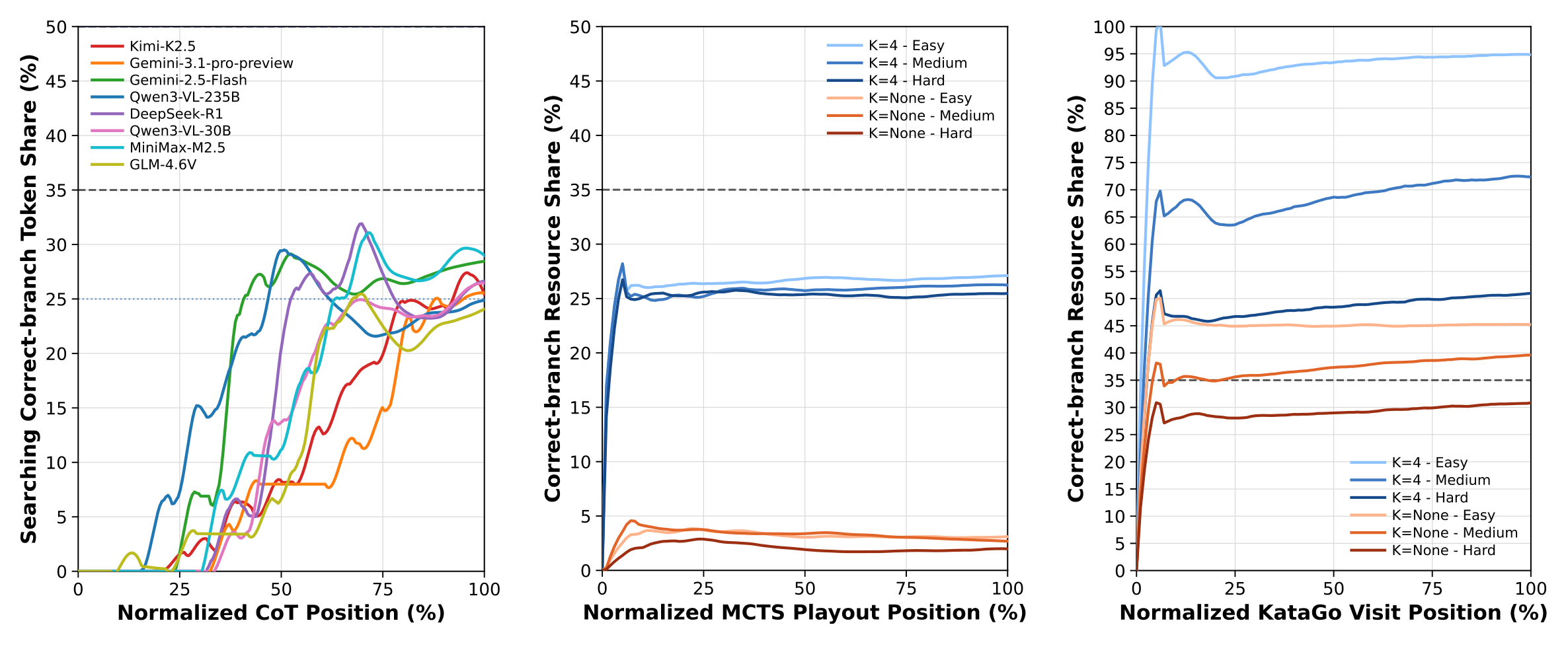}
  \caption{Resource share assigned to the correct branch during search.}
  \label{fig:search_share}
  \vspace{-0.50em}
\end{figure}

\textbf{Dynamics of resource allocation.}
Figure~\ref{fig:search_share} compares how LLM CoT, MCTS, and KataGo allocate resources to the correct branch over normalized search progress; the dashed line marks 35\%.
MCTS and KataGo both form early branch preferences and plateau, but KataGo converges faster and more strongly, especially on Easy and Medium problems.
MCTS stays near random four-choice allocation under $K{=}4$ and lower in open search, while LLM CoT is more scattered and volatile, indicating weaker concentration than explicit search.

\textbf{Domain-specialized model.}
We also test Logos, a Go LLM trained on KataGo game trajectories, but it often continues whole-game play or plays elsewhere instead of resolving the local life-and-death point.
This limitation makes its outputs unsuitable for reasoning-trace analysis and highlights why \tsumego{} targets goal-directed local search rather than move-policy imitation; details appear in the appendix.

\subsection{Bad-Case Analysis and Distribution}
\label{sec:badcase}

Manual inspection of failed open-setting traces identifies failures in candidate-generation knowledge, coordinate grounding, visual perception, board-state maintenance, adversarial verification, instruction following, reasoning loops, and branch management.
These cases indicate a combined bottleneck in candidate generation, local state tracking, adversarial verification, and branch control, rather than resource allocation alone.
Mode distributions and representative cases appear in the appendix.

\section{Conclusion}
\label{sec:conclusion}

We presented \tsumego{}, a process-level benchmark for evaluating Search Efficiency in LLM reasoning. Instead of testing Go-playing strength, it uses life-and-death problems as closed, verifiable, adversarial search spaces; K-Search controls the solution space, and process trees expose resource allocation. Experiments show that current LLMs remain far from stable tsumego solving: stronger models find the correct candidate earlier and sustain productive effort, while failures omit, delay, or abandon the key move, often compounded by state-perception and tactical errors. Search Efficiency is a more effective process signal than token cost alone because it measures how models allocate resources during search. With Token Efficiency as a cost reference, it identifies candidate generation, branch comparison, verification, and backtracking as core search-control bottlenecks. Overall, \tsumego{} turns free-form traces into diagnostic feedback and remains far from saturated, suggesting future work on Search-Efficiency-sensitive training data and planning frameworks.

\bibliography{references}

\clearpage
\appendix

\section{Data Curation and Traceability}
\label{app:data_source}

\tsumego{} is built from traceable tsumego records rather than from free-form puzzle text alone.  Each item is normalized into a board state, side to play, candidate first moves for bounded evaluation, and reference solution lines.  The released benchmark stores the normalized state and solution metadata, while avoiding redistribution of webpage layout, platform-specific presentation, or explanatory prose from the original sources.

\paragraph{Construction process.}
We use SGF as the common intermediate representation.  Source records are first converted into SGF fields for board size, black stones, white stones, side to play, and verified solution moves; the SGF is then converted into the structured problem representation used by all model evaluations.  This keeps the symbolic coordinate state, numeric grid, visual rendering, and answer key tied to the same canonical position.

Each JSON item contains three top-level fields: \texttt{Prompt}, \texttt{Question}, and \texttt{Solution}.  \texttt{Question} stores the symbolic coordinate state, the 19-by-19 grid state, optional visual rendering metadata, and the side to play.  \texttt{Solution} stores the reference first move, the bounded candidate set, all equivalent first moves when present, standard winning lines, variations, and losing candidate lines.  This structure lets every evaluated answer be checked by coordinate normalization rather than by string matching alone.

\paragraph{SGF-to-symbolic conversion.}
SGF points are converted to the benchmark coordinate system by mapping the first SGF character to board columns \texttt{A,B,\ldots,H,J,\ldots,T} and the second SGF character to rows from top to bottom.  Thus an SGF point \texttt{[oe]} maps to \texttt{P15}: \texttt{o} is the fifteenth zero-indexed SGF column, which becomes \texttt{P} after skipping \texttt{I}, and \texttt{e} maps to row $19-4=15$.  The same conversion is applied to setup stones, reference solution lines, candidate options, and model predictions.

\begin{figure*}[t]
  \centering
  \includegraphics[width=.86\textwidth]{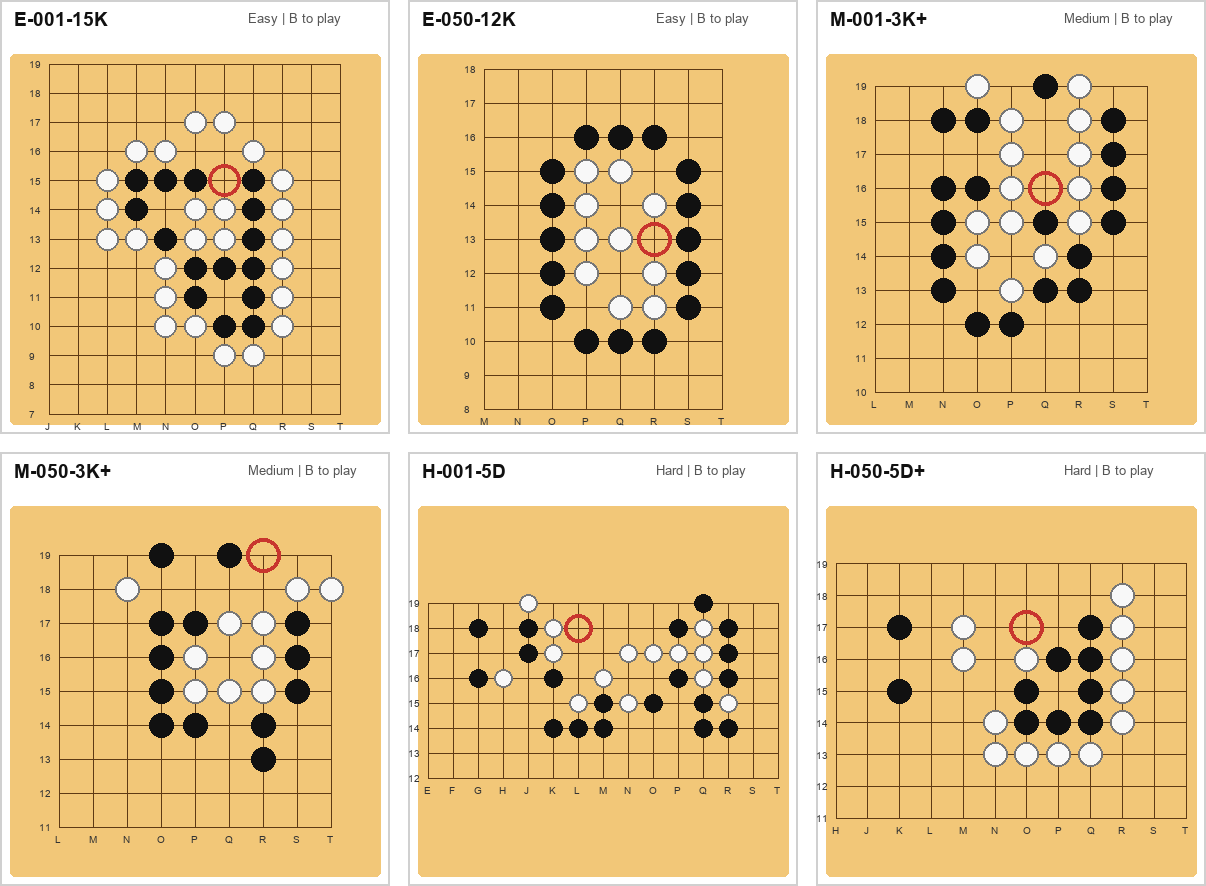}
  \caption{Example normalized tsumego boards sampled from Easy, Medium, and Hard splits.  Labels use benchmark display identifiers with rank tags, and the red circle marks the reference first move.}
  \label{fig:data_example_board}
\end{figure*}

\begin{table}[h]
  \centering
  \caption{Example normalized evaluation interface derived from one SGF problem.}
  \label{tab:normalized_example}
  \scriptsize
  \setlength{\tabcolsep}{3pt}
  \begin{tabular}{lp{.68\columnwidth}}
    \toprule
    Field & Example value \\
    \midrule
    Benchmark record & \texttt{E-001-15K}; Easy split; rank tag \texttt{15K} \\
    SGF state & \texttt{AB[me][pj]\ldots AW[qj][ld]\ldots PL[B]} \\
    Side to play & Black \\
    Symbolic state (Sym) & Black: \texttt{M14,M15,N13,\ldots,Q15}; White: \texttt{L13,L14,\ldots,R15} \\
    Grid state (Grd) & $19{\times}19$ board with 0/1/2 for empty/black/white. \\
    Visual state (Vis) & Rendered board image generated from the same normalized state. \\
    Accepted answers & \texttt{correct\_answer=P15}; \texttt{all\_correct\_answers=\{P15,N14\}} \\
    Candidate options & A: \texttt{P15}; B: \texttt{Q17}; C: \texttt{Q18}; D: \texttt{O18} \\
    Reference lines & Standard: \texttt{[P15]}, \texttt{[N14]}; variations: \texttt{[P15,N14,N13]}, \texttt{[N14,P15,P16]} \\
    Input variants & Sym, Grd, Vis, S+G, and S+V share the same side-to-play, options, and answer key. \\
    \bottomrule
  \end{tabular}
\end{table}

\paragraph{K=4 candidate construction.}
The bounded setting is designed to evaluate first-move discrimination rather than open move generation.  For each problem, the correct first move and all verified equivalent first moves are first identified from the solution record.  Distractor candidates are then drawn from a search-assisted curation pool rather than from random legal moves.  This pool combines non-solution local alternatives recorded during candidate construction, including proposals from non-LLM search references such as KataGo and MCTS/UCT, candidate metadata from the normalized problem record, and geometric locality around the canonical answer.  Candidates that duplicate a verified equivalent answer, land on occupied points, or require global context are removed.  When more than three distractors remain, we rank them by Euclidean board distance $d(m,c^\ast)=\sqrt{(x_m-x_{c^\ast})^2+(y_m-y_{c^\ast})^2}$ to the canonical correct point and choose close alternatives first, because near misses are more diagnostic of local tactical discrimination than arbitrary distant legal moves.

The construction has six steps.  (i) Normalize the SGF board and solution record into the benchmark coordinate system.  (ii) Define the accepted answer set from the canonical solution and all verified equivalent first moves.  (iii) Build a proposal pool from non-solution local alternatives in the construction records, including KataGo and MCTS/UCT proposals.  MCTS/UCT contributes broad candidate coverage, while KataGo contributes more selective candidate and equivalent-answer signals.  (iv) Remove candidates that are occupied, illegal for the side to play, outside the local problem region, or in the accepted answer set.  (v) Use Euclidean distance to prioritize the remaining wrong candidates by locality to the canonical correct point.  (vi) Select the nearest plausible distractors and combine them with the correct answer to form the K=4 set, then randomly shuffle the option labels before evaluation.  This shuffle prevents the verified answer from occupying a fixed A/B/C/D position while preserving the same coordinate-level answer key.  The resulting sampled problems are then checked by human review and independent search traces.  Search references are therefore used to propose plausible wrong moves, and distance is used to choose local near misses; neither defines correctness, which always comes from the verified solution record.

\begin{algorithm}[t]
\caption{Bounded K=4 Candidate Construction}
\label{alg:k4_construction}
\begin{algorithmic}[1]
\REQUIRE Normalized board $B$, side to play $s$, canonical answer $c^\ast$, verified equivalent answers $E$, search-assisted proposal records $S_{\mathrm{KG}},S_{\mathrm{MCTS}}$
\ENSURE Four-option set $O$
\STATE $A \leftarrow \{c^\ast\} \cup E$ \COMMENT{accepted first-move answer set}
\STATE $P \leftarrow \{m: m \textrm{ is a non-solution local proposal in } S_{\mathrm{KG}} \cup S_{\mathrm{MCTS}}\}$
\STATE $P \leftarrow P \setminus A$
\STATE remove occupied, illegal, non-local, and globally context-dependent moves from $P$
\STATE rank each $m \in P$ by Euclidean distance to $c^\ast$
\STATE $D \leftarrow$ the nearest three remaining moves
\STATE $O \leftarrow \mathrm{Shuffle}(\{c^\ast\}\cup D)$
\STATE \textbf{return} $O$
\end{algorithmic}
\end{algorithm}

A post-hoc audit of independent shallow-search traces supports this curation rule.  Table~\ref{tab:k4_search_audit} summarizes the 300-problem bounded-candidate trace-audit subset, sampled as 100 problems from each evaluated difficulty tier.  This subset is half of the 600-problem main evaluation and is the portion for which complete open-root and restricted-root MCTS/UCT and KataGo root-distribution traces are available.  The audit is deliberately separate from the construction step above: it does not define the options, but applies the same coverage test to independent root-distribution traces.  Coverage is therefore not expected to be 100\%.  The audit has two purposes.  First, it checks that selected non-correct options are usually search-visible rather than arbitrary legal points.  Second, it checks that additional equivalent correct moves are recognized as accepted answers rather than counted as distractors.  For each selected option, we scan budgets up to 30 and count it as covered if it receives a nonzero root visit.  The 881 non-correct option entries are obtained by counting all final K=4 options in the 300 sampled problems and excluding verified correct or equivalent-correct entries: $300\times4=1200$ total option entries, $319$ accepted-answer entries, and $1200-319=881$ non-correct entries.

Open-root coverage is a plausibility check, not a correctness signal.  In an open-root run, KataGo or MCTS/UCT searches from the original board without being told the four benchmark options, so a distractor covered by open-root search is a move that the search process naturally considered among legal root moves.  Restricted-root coverage answers a different question: after the benchmark fixes the final K=4 interface, does the search process allocate visits to those listed options?  We therefore use open-root coverage to test whether distractors are naturally plausible, and restricted-root coverage to test whether the final option set remains meaningful under the same answer interface used by LLMs.

\begin{table}[t]
  \centering
  \caption{Audit of K=4 candidate construction using independent shallow-search traces for the 300-problem trace-audit subset.  The 881 non-correct option entries equal $300\times4$ final option entries minus 319 verified correct or equivalent-correct entries.  A candidate is counted as covered when it receives a nonzero root visit within budget 0--30.  Open-root runs search all legal root moves; restricted-root runs search only the final K=4 option set.}
  \label{tab:k4_search_audit}
  \scriptsize
  \setlength{\tabcolsep}{3pt}
  \begin{tabular}{lrr}
    \toprule
    Audit quantity & Count & Rate \\
    \midrule
    Problems in curation subset & 300 & -- \\
    Final option entries & 1200 & -- \\
    Accepted-answer entries & 319 & -- \\
    Non-correct option entries & 881 & -- \\
    Covered by open-root MCTS/UCT & 629/881 & 71.4\% \\
    Covered by open-root KataGo & 210/881 & 23.8\% \\
    Covered by either open-root search & 681/881 & 77.3\% \\
    Covered by restricted-root MCTS/UCT & 851/881 & 96.6\% \\
    Covered by restricted-root KataGo & 485/881 & 55.1\% \\
    Extra equivalent correct moves & 38 & -- \\
    Equiv. correct visited by open-root MCTS/UCT & 26/38 & 68.4\% \\
    Equiv. correct visited by open-root KataGo & 23/38 & 60.5\% \\
    \bottomrule
  \end{tabular}
\end{table}

The table shows that 77.3\% of the final non-correct options are independently encountered by at least one open-root search reference within the shallow budget.  MCTS/UCT contributes broader early root coverage, while KataGo is more selective; this difference is useful because it prevents the option set from reflecting a single search policy.  These visit counts are used as priority and plausibility signals, not as correctness thresholds.  The restricted-root columns provide a second check: once the final four candidates are fixed, many non-correct options receive visits under the same answer interface used for model evaluation.  The equivalent-correct rows also show why answer-set filtering must precede distractor selection: many verified equivalent answers are attractive to search references, so using search proposals alone would incorrectly label some valid first moves as distractors.

\begin{table}[t]
  \centering
  \caption{Example K=4 evidence for benchmark problem \texttt{E-001-15K}.  Rows A--D are the final answer options, while \texttt{N14} is shown as a rejected proposal.  Distance is Euclidean distance to the canonical answer \texttt{P15}; first-seen columns report the earliest independent audit budget, at most 30, where the move receives a nonzero root visit.  \texttt{N14} is search-visible but removed because it is a verified equivalent correct answer.}
  \label{tab:k4_example_distance}
  \scriptsize
  \setlength{\tabcolsep}{2pt}
  \begin{tabular}{lllrrrrr}
    \toprule
    Option & Coord. & Label & Dist. & MCTS-O & KG-O & MCTS-4 & KG-4 \\
    \midrule
    A & P15 & correct & 0.00 & 21 & 2 & 1 & 2 \\
    B & Q17 & distractor & 2.24 & 11 & -- & 2 & 30 \\
    C & Q18 & distractor & 3.16 & 7 & -- & 3 & -- \\
    D & O18 & distractor & 3.16 & 5 & -- & 4 & -- \\
    -- & N14 & equiv. correct & 2.24 & 27 & 4 & -- & -- \\
    \bottomrule
  \end{tabular}
\end{table}

Here ``O'' denotes open-root search over legal root moves, and ``4'' denotes restricted-root search over the final K=4 option set.  The example illustrates the full curation-and-audit chain.  The final distractors are local near misses around the canonical answer and appear in independent MCTS audit traces, so they are not random legal points.  At the same time, \texttt{N14} is also search-visible and geometrically close, but it is filtered out before final option selection because the verified solution record marks it as an equivalent correct first move.

\paragraph{Evaluation prompt example.}
For the same problem, the symbolic prompt presents the state as normalized coordinates rather than as a natural-language puzzle description: ``Black stones: \texttt{M14, M15, N13, \ldots, Q15}; White stones: \texttt{L13, L14, \ldots, R15}; side to play: Black; choose from A: \texttt{P15}, B: \texttt{Q17}, C: \texttt{Q18}, D: \texttt{O18}.''  The final answer is scored by mapping the option letter back to a normalized coordinate and checking whether it belongs to the verified correct-answer set.

\paragraph{Filtering criteria.}
We retain a problem only when the local life-and-death objective is well defined, the first move is unique or belongs to a finite equivalence class, the candidate set is legal under coordinate normalization, and the reference line is not dependent on unresolved global ko or whole-board context.  Ambiguous problems, records with inconsistent side-to-play metadata, positions whose candidate labels cannot be mapped back to legal board coordinates, and K=4 sets containing another unverified equivalent answer are removed before evaluation.

\paragraph{Normalization checks.}
Every retained item is checked at three levels.  First, all coordinates must use the same 19-by-19 convention with the column \texttt{I} omitted.  Second, the symbolic coordinate state and grid state must agree after color, row, and column normalization.  Third, each candidate first move and each reference first move must map to an empty legal coordinate for the stated side to play.  These checks make answer evaluation independent of the surface text used to present a problem.

\paragraph{Traceability.}
The benchmark keeps source-level identifiers, rank tags, candidate labels, and reference-line metadata for auditability, but evaluation uses the normalized board state.  This separation is important for reproducibility.  A model can be evaluated from the released structured problem alone, while the construction record still allows us to inspect why a puzzle was included, how its candidates were produced, and whether an apparent model error is actually a coordinate or metadata issue.

\paragraph{Why the bounded task is evaluable.}
The K=4 setting is intentionally conservative.  Five human reviewers, assisted by KataGo line checking, review the first-move answer set and the distractor set for 300 sampled problems, corresponding to half of the 600-problem main evaluation.  The reviewers self-rate as two stronger amateur Go players, two intermediate amateur players, and one beginner amateur player; each reviewer spends approximately 12 hours on the review, for about 60 reviewer-hours before joint adjudication.  The review asks whether the listed correct moves are all acceptable first-move tesuji, whether any distractor is actually an equivalent solution, and whether every option is a legal local move under the normalized state.  Accuracy is then defined only at the first move: a model is correct if its selected coordinate is in the verified correct-answer set.  This does not claim to certify every possible continuation of an open game; it certifies the local four-choice eye point used by the benchmark.

This bounded evaluation is already non-saturated.  Even on the easiest K=4 tier, most evaluated LLMs remain well below perfect accuracy in the main table.  Therefore, although K=4 is simpler than open search, it still provides a meaningful first layer of measurement: it tests whether the model can identify the vital point among locally plausible alternatives before the benchmark asks it to generate candidates from the full board.

\begin{table}[t]
  \centering
  \caption{Composition audit for the 300-problem bounded-candidate curation subset used for structural checks.  This audit subset is separate from the 600-problem main evaluation set.}
  \label{tab:dataset_audit}
  \small
  \setlength{\tabcolsep}{4pt}
  \begin{tabular}{lrrrrr}
    \toprule
    Split & N & Options & B/W & Std. len. & Var. len. \\
    \midrule
    Easy & 100 & 4 & 99/1 & 2.77 & 3.84 \\
    Medium & 100 & 4 & 68/32 & 6.56 & 6.86 \\
    Hard & 100 & 4 & 85/15 & 11.92 & 10.14 \\
    \bottomrule
  \end{tabular}
\end{table}

The split statistics in Table~\ref{tab:dataset_audit} show the intended difficulty progression: harder tiers have longer standard solution lines and longer failure/variation lines.  This pattern is important because it verifies that the split is not merely relabeled by rank; the reference solution structure itself becomes longer and more branching as difficulty increases.

\section{Equivalence-Transform Robustness}
\label{app:robustness}

The benchmark is designed so that tactical content is invariant to board symmetries and color relabeling.  We therefore define a transform suite for robustness evaluation: 90/180/270 degree rotation, horizontal reflection, vertical reflection, color inversion with side-to-play flip, and coordinate relabeling.  This test is not a proof that no benchmark position appears in pretraining data.  Instead, it is a surface-memorization stress test: the local tsumego answer is preserved while coordinates, colors, or geometry are changed.  A model that relies on memorized board layouts or coordinate shortcuts should become less consistent under these equivalent views.

\paragraph{Transform construction audit.}
Before using the transforms for model evaluation, we audit whether each transformed problem preserves a valid coordinate system and maps the reference answer into the transformed candidate set.  The construction audit covers 300 benchmark items and passes for all tested transforms: rotation, horizontal/vertical reflection, and color inversion.  This check ensures that a model failure is not caused by an invalid transformed instance.

\paragraph{Robustness protocol.}
We evaluate transformation invariance with a controlled answer-only protocol at temperature 0.  The prompt asks for a single option letter, so this probe is used to study presentation sensitivity rather than to replace the main benchmark accuracy estimates.  For each evaluated model, we sample 10 problems from the benchmark with difficulty stratification across Easy, Medium, and Hard items.  Each sampled item is evaluated under the original presentation and three answer-preserving transformations.  Table~\ref{tab:transform_audit} reports the average result across models.

Accuracy is computed against the answer after transformation.  Consistency is computed by mapping each transformed prediction back to the original coordinate system and comparing it with the same model's prediction on the original presentation of the same problem.  The latter is the primary robustness quantity: a model can be inaccurate in this terse setting yet still be invariant, or accurate on some transformed prompts while changing its underlying prediction.

\begin{table}[t]
  \centering
  \caption{Average equivalence-transform robustness over evaluated models.  For each model, 10 benchmark problems are sampled with difficulty stratification.  Acc. is accuracy under the corresponding surface form; Consist. is inverse-mapped agreement with the same model's original prediction.}
  \label{tab:transform_audit}
  \small
  \setlength{\tabcolsep}{3pt}
  \begin{tabular}{lrrr}
    \toprule
    Transform & Problems/model & Acc. & Consist. \\
    \midrule
    Original & 10 & 26.7 & 100.0 \\
    Rotate 180 & 10 & 40.0 & 60.0 \\
    Mirror horizontal & 10 & 31.1 & 64.4 \\
    Color inversion & 10 & 28.9 & 77.8 \\
    \bottomrule
  \end{tabular}
\end{table}

\paragraph{Interpretation.}
The aggregate results show that answer-preserving surface changes can substantially alter model behavior.  Geometric transforms reduce consistency, with 60.0\% under 180-degree rotation and 64.4\% under horizontal reflection, suggesting that spatial normalization remains imperfect even when the problem is presented symbolically.  Color inversion is less disruptive but still below perfect consistency at 77.8\%, indicating sensitivity to color and side-to-play relabeling.  These results do not prove absence of training-set exposure; they support the more limited claim used in the main paper, namely that surface-equivalent perturbations are a useful robustness control for memorization-like shortcuts and coordinate sensitivity.

\section{Process-Tree Extraction and Validation}
\label{app:extraction_validation}

The main paper evaluates visible reasoning by converting free-form responses into process search trees.  We use a compact node schema with three node types: \texttt{root}, \texttt{move}, and \texttt{judgment}.  Candidate proposal, adversarial reply, variation continuation, and branch switching are metric-level roles derived from node depth, parent links, move side, and temporal order.  The root is the initial board state; first-level move children are candidate first moves whenever they can be recovered.

\paragraph{Extractor protocol.}
We use Gemini-2.5-Pro as the primary visible-trace extractor because it gives stable structured outputs on long free-form reasoning while preserving enough local Go terminology to avoid excessive rule-based preprocessing.  The prompt below is a condensed English version of the extraction instructions used in evaluation.  The extractor is not given the reference answer as a target to imitate.  It receives the model's visible reasoning text and a schema.  The prompt asks the extractor to preserve the temporal order of the original response, identify explicitly analyzed moves, attach follow-up moves to the active branch, and mark terminal judgments as \texttt{win}, \texttt{lose}, or \texttt{other}.  A structural validation step then enforces exactly one root, valid parent pointers, move labels with side and coordinate when recoverable, no orphaned nodes, and monotone node order.  Repeated sibling moves are merged when they clearly refer to the same move under the same parent; incomplete branches are closed with an \texttt{other} judgment instead of being deleted.

\paragraph{Extractor prompt.}
The extraction prompt first defines the rules, then includes a compact structured example showing the required node format, and finally appends the model reasoning trace to be parsed.  The version below condenses the instruction portion while preserving the metric-relevant constraints:
\begin{quote}\small
\textbf{Task.} Convert the visible reasoning trace into a structured timeline search tree for a Go life-and-death problem.  Read the model's thinking/reasoning text and output a JSON tree describing which move branches were explicitly considered, how each branch was expanded, and what conclusion was reached.

\textbf{Node types.} Use only three node types.  \texttt{root} appears once with \texttt{id=0} and \texttt{parent=null}.  \texttt{move} represents an explicitly analyzed move hypothesis and must include \texttt{side} in \{\texttt{B},\texttt{W}\} and \texttt{move\_pos}.  \texttt{judgment} terminates a branch and must be a leaf with \texttt{polarity} in \{\texttt{win},\texttt{lose},\texttt{other}\}.

\textbf{Extraction scope.} Extract necessary structure only: candidate moves considered by the model, explicitly expanded follow-up moves, and final branch judgments.  Do not convert background board description, repeated restatement, global summaries, or comparison prose into nodes.  If a move is merely mentioned but not analyzed as a branch, do not create a node for it.

\textbf{Timeline and tree rules.} Node ids must strictly increase in the order the response introduces them.  The root may have only \texttt{move} children.  Every non-leaf node except the root must be a \texttt{move}.  Every leaf must be a \texttt{judgment}.  If the same move appears under different parents or different search branches, create a new node; if the same move is repeated under the same parent, reuse the existing node.  Do not invent intermediate moves to complete a line.

\textbf{Judgment rules.} Add \texttt{judgment} only when a branch terminates.  Do not turn intermediate statements such as ``has one liberty'', ``connects'', ``preliminary comparison'', or ``think further'' into judgments if the response continues the line.  Use \texttt{win} for a branch judged feasible, alive, killing, optimal, or successful; \texttt{lose} for a branch judged failed, captured, refuted, or inferior; and \texttt{other} for pruned, unclear, interrupted, or unresolved branches.

\textbf{Output.} Return only a JSON object with a \texttt{nodes} array.  Each node includes \texttt{id}, \texttt{label}, \texttt{type}, \texttt{parent}, and, for moves, \texttt{side} and \texttt{move\_pos}; for judgments, include \texttt{polarity}.  Do not output explanations or Markdown.
\end{quote}

The structured example has the following form:
\begin{quote}\scriptsize
\begin{verbatim}
{
  "nodes": [
    {
      "id": 0,
      "label": "Initial board state",
      "type": "root",
      "parent": null
    },
    {
      "id": 1,
      "label": "B:T14",
      "type": "move",
      "parent": 0,
      "side": "B",
      "move_pos": "T14"
    },
    {
      "id": 2,
      "label": "W:S13",
      "type": "move",
      "parent": 1,
      "side": "W",
      "move_pos": "S13"
    },
    {
      "id": 3,
      "label": "Black lives",
      "type": "judgment",
      "parent": 2,
      "polarity": "win"
    }
  ]
}
\end{verbatim}
\end{quote}

\begin{table}[t]
  \centering
  \caption{Node types used by the visible process-tree extractor.}
  \label{tab:node_types}
  \small
  \setlength{\tabcolsep}{3pt}
  \begin{tabular}{lp{.68\columnwidth}}
    \toprule
    Node type & Extraction rule \\
    \midrule
    Root & Initial board state and side to play. \\
    Move & Explicitly analyzed move or move hypothesis; first-level moves are candidate first moves. \\
    Judgment & Branch-ending win, loss, refutation, success, uncertainty, or pruning claim. \\
    Derived roles & Candidate, reply, continuation, and branch switch are inferred from depth, side, parent, and time order. \\
    \bottomrule
  \end{tabular}
\end{table}

\paragraph{Validation protocol.}
Human validation samples across model, difficulty, modality, candidate-space condition, and correctness strata.  The validation set contains 300 sampled problem responses, matching the sample size referenced in the main paper.  Two human experts independently inspect the Gemini-2.5-Pro extraction against the original visible response.  The experts check five metric-critical properties: first-level candidate recovery, candidate appearance order, parent-child branch assignment, terminal polarity, and whether the resulting tree preserves the values needed for SWR, SFH, SCR, TMD, TMF, TNC, IPN, and ITT.  The reported 93--98\% range comes from this independent cross-validation: most disagreements are semantic boundary cases rather than structurally invalid trees, such as a move being briefly mentioned but not developed into a branch, a branch being revisited without naming the first move again, a pronoun such as ``this line'' having multiple possible antecedents, or a summary-only trace compressing several branches into one sentence.  The experts are assisted by the normalized board state and KataGo line checking when a move label or terminal Go judgment is ambiguous, but the adjudication target remains the visible text, not the engine's preferred solution.  Disagreements are resolved by joint adjudication.  Ambiguous proprietary summaries are not expanded into hidden internal trees; they are marked as summary-derived traces.

\begin{table}[t]
  \centering
  \caption{Human cross-validation of Gemini-2.5-Pro process-tree extraction on 300 sampled problem responses.  Rates report whether the automatic extraction passes the human check for each metric-critical property before joint adjudication; the 93--98\% range mainly reflects semantic boundary ambiguity in free-form traces.}
  \label{tab:extractor_validation}
  \small
  \setlength{\tabcolsep}{3pt}
  \begin{tabular}{p{.42\columnwidth}cp{.34\columnwidth}}
    \toprule
    Validation property & Pass rate & Main risk when failed \\
    \midrule
    First-level candidate set & 98\% & brief mention vs.\ analyzed branch \\
    Candidate order / SFH rank & 97\% & late restatement of an earlier branch \\
    Parent-child branch edges & 95\% & pronoun or ``this line'' ambiguity \\
    Terminal polarity & 93\% & hedged or summary-level life/death judgment \\
    Metric-consistent tree & 96\% & merged or split repeated branch \\
    \bottomrule
  \end{tabular}
\end{table}

\paragraph{Validation criteria.}
We judge extraction quality at the level required by the metrics rather than by exact natural-language paraphrase.  A tree is considered metric-consistent when it preserves (i) whether the correct candidate appears, (ii) the order in which first-level candidates appear, (iii) the allocation of explored nodes to correct and wrong first-move branches, (iv) the terminal polarity of each explicitly evaluated branch, and (v) the observable token span used to compute scale metrics.  This criterion is stricter than answer extraction but more stable than requiring annotators to agree on every intermediate phrase in a long free-form trace.

\paragraph{Observed extraction risks.}
The main failure modes of extraction are implicit candidate references, coordinate aliases, summary-only proprietary reasoning, and long responses that revisit a branch without explicitly naming the first move.  We handle these cases conservatively: ambiguous branches remain marked as unclear, summary-only traces are not expanded into hidden internal search, and answer metrics are computed separately from process metrics.

\paragraph{Why tree validation is metric-specific.}
The goal is not to reconstruct private cognition, but to measure the organization of visible evidence.  Therefore the extractor is allowed to ignore rhetorical filler, self-corrections that do not introduce a new board state, and repeated restatements of an already represented branch.  It is not allowed to add a candidate that is only implied by the reference answer, to move an opponent reply under a more convenient parent, or to convert an uncertain branch into a terminal win/loss.  These constraints keep SWR, SFH, CCER, and related metrics tied to the text actually shown by the model.

\section{Metric Definitions and Weight Sensitivity}
\label{app:metric_formulas}

\paragraph{Answer Accuracy.}
For a benchmark set $\mathcal{D}=\{x_i\}_{i=1}^n$, let $c_i^*$ be the reference first move and $\hat{c}_i$ be the first move selected by the model after coordinate normalization.  We compute answer accuracy as
\begin{equation}
  \mathrm{Acc} = \frac{1}{n}\sum_{i=1}^{n}\mathbb{1}[\hat{c}_i = c_i^*].
\end{equation}

\paragraph{SearchE.}
The main table reports a 0--100 structural-efficiency score:
\begin{equation}
  \mathrm{SearchE}
  =
  100\cdot\bigl(0.5(1-\swr)+0.3\,\sfh+0.2(1-\scr)\bigr).
\end{equation}
It aggregates wrong-branch waste, early first exploration of the correct candidate, and the search rank of that candidate.

\paragraph{TokenE.}
Let $A=100\cdot\mathrm{Acc}$ be accuracy in percentage points and let $\itt_{\mathrm{raw}}$ be the raw number of observable thinking tokens.  We report token efficiency as
\begin{equation}
  \mathrm{TokenE}
  =
  100\cdot \frac{A}{A+\itt_{\mathrm{raw}}/1000}.
\end{equation}
TokenE is a cost reference for LLM traces and is undefined for search baselines, which spend playouts or visits rather than language tokens.

\begin{table*}[t]
  \centering
  \caption{Complete metric inventory used in the main and appendix result tables.  Search-organization metrics explain where visible reasoning effort is allocated; trajectory metrics describe extracted tree topology; scale metrics describe observable text budget.}
  \label{tab:metric_inventory}
  \small
  \setlength{\tabcolsep}{4pt}
  \begin{tabular}{>{\raggedright\arraybackslash}p{.12\textwidth}
                  >{\raggedright\arraybackslash}p{.18\textwidth}
                  >{\raggedright\arraybackslash}p{.12\textwidth}
                  >{\raggedright\arraybackslash}p{.50\textwidth}}
    \toprule
    Metric & Category & Direction & Definition / interpretation \\
    \midrule
    Acc. & Answer & $\uparrow$ & First-move hit rate after coordinate normalization. \\
    SearchE & Composite & $\uparrow$ & Weighted score from SWR, SFH, and SCR. \\
    TokenE & Composite & $\uparrow$ & Accuracy normalized by observable token cost. \\
    $\swr$ & Search organization & $\downarrow$ & Fraction of explored tree nodes under wrong first-move branches. \\
    $\sfh$ & Search organization & $\uparrow$ & Indicator that the correct first-move branch is explored first. \\
    CCER & Search organization & $\downarrow$ & Rank at which the correct first-move branch first appears; $\infty$ if absent. \\
    $\scr$ & Search organization & $\downarrow$ & Normalized correct-candidate rank derived from CCER and SBC. \\
    $\ssc$ & Search organization & context & Number of switches between first-level candidate branches. \\
    $\sbc$ & Search organization & context & Number of first-level candidate branches considered. \\
    $\tmd$ & Trajectory & context & Maximum root-to-leaf depth of the extracted process tree. \\
    $\tmf$ & Trajectory & context & Maximum number of children of any single tree node. \\
    $\tnc$ & Trajectory & context & Number of nodes in the extracted process tree. \\
    $\itt$ & Scale & $\downarrow$ & Observable reasoning-token count; reported in thousands in compact tables. \\
    $\ipn$ & Scale & $\downarrow$ & Observable reasoning tokens per extracted tree node. \\
    \bottomrule
  \end{tabular}
\end{table*}

\paragraph{Search-tree notation.}
For one response, let $T=(V,E,r,\tau)$ be the extracted process tree, where $r$ is the root and $\tau$ orders nodes by their appearance in the reasoning trace.  Let $C=\{c_1,\ldots,c_m\}$ be first-level candidate moves, and let $c^*$ denote the correct first move.  For any node $v$, $\mathrm{owner}(v)$ denotes the first-level candidate whose subtree contains $v$.  Let $u_1,\ldots,u_L$ be the non-root nodes sorted by $\tau$.

\paragraph{Search metrics.}
\begin{equation}
\begin{aligned}
  \mathrm{CCER}
  &= \min\{j:\mathrm{owner}(u_j)=c^*\},\\
  \mathrm{CCER}
  &= \infty
  \quad \textrm{if } c^* \textrm{ is never expanded.}
\end{aligned}
\end{equation}
\begin{equation}
  \swr =
  \frac{\sum_{c \in C,\, c \neq c^*} |\mathrm{subtree}(c)|}
       {\sum_{c \in C} |\mathrm{subtree}(c)|}.
\end{equation}
\begin{equation}
  \sfh = \mathbb{1}[\mathrm{CCER}=1].
\end{equation}
\begin{equation}
  \scr =
  \begin{cases}
    (\mathrm{CCER}-1)/\sbc, & \mathrm{CCER}<\infty,\\
    1, & \mathrm{CCER}=\infty.
  \end{cases}
\end{equation}
\begin{equation}
\begin{aligned}
  \ssc
  &= \sum_{\ell=2}^{L}
  \mathbb{1}[
  \mathrm{owner}(u_{\ell}) \neq \mathrm{owner}(u_{\ell-1})],\\
  \sbc
  &= |C|.
\end{aligned}
\end{equation}
\begin{equation}
\begin{aligned}
  \itt
  &= \textrm{number of observable reasoning tokens},\\
  \ipn
  &= \itt / |V|.
\end{aligned}
\end{equation}
\begin{equation}
\begin{aligned}
  \tmd &= \max_{v \in V}\mathrm{dist}(r,v),\\
  \tnc &= |V|,\\
  \tmf &= \max_{v \in V}|\mathrm{children}(v)|.
\end{aligned}
\end{equation}
The formulas above use fractional SWR, SFH, and SCR values in $[0,1]$.  All result tables report these three metrics on a 0--100 scale for consistency with the main table captions and SearchE.

\paragraph{Complete result tables.}
Table~\ref{tab:appendix_complete_results} provides the full aggregate diagnostic layout: it reports accuracy, search-organization metrics $\swr$, $\sfh$, $\scr$, $\ssc$, and $\sbc$, trajectory metrics $\tmd$, $\tmf$, and $\tnc$, and scale metrics $\ipn$ and $\itt(k)$ for LLMs, proprietary systems, and search baselines.  It omits SearchE and TokenE because those composite summaries already appear in the main table.  Table~\ref{tab:appendix_full_llm_results} gives the complementary per-modality LLM accuracy breakdown, while trace-level diagnostics remain in Table~\ref{tab:appendix_complete_results}.

\begin{table*}[p]
  \caption{Complete diagnostic results with the original non-composite metric columns.  Each metric cell uses the format $K{=}4\,|\,K{=}\mathrm{None}$.  Accuracy averages available input modalities; all trace metrics are computed under Symbolic input.  SWR, SFH, and SCR are reported on a 0--100 scale without percent signs.  Scale follows the main table: for LLMs, ITT is reported in thousand tokens; for search baselines, TNC is reported in thousands and the final column reports playout/visit budget rather than tokens.  Gray proprietary values are summary-derived and are not directly comparable to full visible traces.}
  \label{tab:appendix_complete_results}
  \centering
  \scriptsize
  \setlength{\tabcolsep}{1.6pt}
  \renewcommand{\arraystretch}{1.02}
  \resizebox{\textwidth}{!}{%
  \begin{tabular}{ll c|ccccc|ccc|cc}
    \toprule
    \textbf{Model} & \textbf{Diff.} & \textbf{Acc} & \multicolumn{5}{c|}{\textbf{Search Organization}} & \multicolumn{3}{c|}{\textbf{Trajectory}} & \multicolumn{2}{c}{\textbf{Scale}} \\
    \cmidrule(lr){4-8} \cmidrule(lr){9-11} \cmidrule(lr){12-13}
    & & & $\mathit{SWR}\downarrow$ & $\mathit{SFH}\uparrow$ & $\mathit{SCR}\downarrow$ & $\mathit{SSC}$ & $\mathit{SBC}$ & $\mathit{TMD}$ & $\mathit{TMF}$ & $\mathit{TNC}$ & $\mathit{IPN}\downarrow$ & $\mathit{ITT}(k)\downarrow$ \\
    \midrule
    \multicolumn{13}{l}{\textbf{\textit{Reasoning Models}}} \\
    \multirow{3}{*}{Kimi-K2.5} & Easy & $52.0\,|\,27.8$ & $66.8\,|\,82.3$ & $31.6\,|\,18.4$ & $30.7\,|\,39.8$ & $4.2\,|\,6.5$ & $4.1\,|\,6.4$ & $4.6\,|\,4.8$ & $2.3\,|\,2.4$ & $20.5\,|\,24.7$ & $1269\,|\,1176$ & $24.3\,|\,31.1$ \\
     & Med & $28.4\,|\,11.0$ & $75.2\,|\,92.6$ & $23.2\,|\,8.2$ & $35.5\,|\,70.9$ & $4.3\,|\,6.8$ & $4.1\,|\,6.8$ & $4.7\,|\,5.3$ & $2.1\,|\,2.5$ & $21.2\,|\,26.2$ & $1258\,|\,1014$ & $22.3\,|\,28.7$ \\
     & Hard & $34.0\,|\,4.4$ & $71.8\,|\,92.4$ & $20.2\,|\,5.4$ & $36.3\,|\,74.9$ & $4.5\,|\,7.8$ & $4.0\,|\,7.6$ & $5.0\,|\,4.9$ & $2.3\,|\,2.2$ & $19.6\,|\,26.8$ & $1213\,|\,1197$ & $21.3\,|\,31.8$ \\
    \addlinespace[1.3pt]
    \multirow{3}{*}{\shortstack[l]{Qwen3-VL-235B-\\Thinking}} & Easy & $40.0\,|\,15.8$ & $69.9\,|\,84.2$ & $33.7\,|\,7.1$ & $30.5\,|\,52.5$ & $3.6\,|\,4.5$ & $4.1\,|\,5.3$ & $5.1\,|\,5.1$ & $2.2\,|\,2.3$ & $24.4\,|\,27.3$ & $854\,|\,745$ & $18.3\,|\,17.9$ \\
     & Med & $25.0\,|\,8.8$ & $75.1\,|\,92.4$ & $23.5\,|\,12.6$ & $37.0\,|\,73.4$ & $3.3\,|\,4.9$ & $4.0\,|\,5.6$ & $5.4\,|\,4.9$ & $2.0\,|\,2.1$ & $22.9\,|\,23.7$ & $815\,|\,858$ & $17.8\,|\,18.5$ \\
     & Hard & $32.0\,|\,7.2$ & $73.8\,|\,94.2$ & $26.5\,|\,3.2$ & $34.4\,|\,82.6$ & $3.4\,|\,5.1$ & $4.0\,|\,5.9$ & $5.4\,|\,5.0$ & $2.2\,|\,2.1$ & $23.2\,|\,25.0$ & $901\,|\,794$ & $19.2\,|\,18.9$ \\
    \addlinespace[1.3pt]
    \multirow{3}{*}{\shortstack[l]{Qwen3-VL-30B-\\Thinking}} & Easy & $34.2\,|\,10.0$ & $73.0\,|\,90.0$ & $30.0\,|\,7.0$ & $35.5\,|\,41.7$ & $3.7\,|\,4.7$ & $4.0\,|\,5.2$ & $3.9\,|\,3.7$ & $1.5\,|\,1.4$ & $15.6\,|\,17.5$ & $1216\,|\,1264$ & $17.9\,|\,19.3$ \\
     & Med & $23.0\,|\,4.6$ & $76.0\,|\,96.0$ & $24.0\,|\,5.0$ & $36.1\,|\,45.1$ & $3.8\,|\,4.9$ & $4.1\,|\,5.5$ & $3.7\,|\,3.6$ & $1.4\,|\,1.2$ & $14.7\,|\,17.3$ & $1285\,|\,1304$ & $18.0\,|\,19.2$ \\
     & Hard & $30.2\,|\,3.6$ & $74.0\,|\,97.0$ & $24.0\,|\,2.0$ & $36.8\,|\,57.8$ & $3.9\,|\,5.5$ & $4.0\,|\,5.8$ & $3.8\,|\,3.4$ & $1.5\,|\,1.2$ & $15.2\,|\,17.7$ & $1259\,|\,1256$ & $18.0\,|\,19.2$ \\
    \addlinespace[1.3pt]
    \multirow{3}{*}{\shortstack[l]{MiniMax-M2.5\\{\scriptsize (Non-VL)}}} & Easy & $37.7\,|\,13.3$ & $70.7\,|\,87.0$ & $28.0\,|\,7.1$ & $35.3\,|\,62.9$ & $4.0\,|\,5.7$ & $4.0\,|\,5.6$ & $3.5\,|\,3.5$ & $1.7\,|\,1.6$ & $15.0\,|\,18.3$ & $827\,|\,687$ & $11.4\,|\,11.1$ \\
     & Med & $22.0\,|\,7.3$ & $75.3\,|\,93.1$ & $18.0\,|\,9.1$ & $38.2\,|\,80.8$ & $3.7\,|\,5.3$ & $4.0\,|\,5.7$ & $3.1\,|\,3.5$ & $1.2\,|\,1.3$ & $12.6\,|\,17.3$ & $540\,|\,740$ & $5.2\,|\,11.6$ \\
     & Hard & $26.3\,|\,6.3$ & $74.2\,|\,94.2$ & $21.0\,|\,4.5$ & $38.7\,|\,84.4$ & $4.3\,|\,6.3$ & $4.0\,|\,6.6$ & $3.3\,|\,3.3$ & $1.5\,|\,1.4$ & $14.5\,|\,12.9$ & $734\,|\,722$ & $9.5\,|\,12.5$ \\
    \addlinespace[1.3pt]
    \multirow{3}{*}{\shortstack[l]{DeepSeek-R1-\\0528{\scriptsize (Non-VL)}}} & Easy & $32.3\,|\,17.0$ & $69.2\,|\,83.5$ & $30.3\,|\,5.3$ & $30.7\,|\,53.5$ & $3.3\,|\,4.8$ & $4.0\,|\,5.6$ & $4.6\,|\,4.5$ & $2.0\,|\,2.1$ & $19.5\,|\,23.3$ & $743\,|\,676$ & $13.1\,|\,14.6$ \\
     & Med & $26.3\,|\,8.3$ & $75.5\,|\,94.1$ & $20.0\,|\,3.0$ & $40.2\,|\,81.2$ & $3.5\,|\,5.1$ & $4.0\,|\,5.7$ & $4.6\,|\,4.3$ & $2.1\,|\,2.1$ & $21.2\,|\,22.6$ & $547\,|\,680$ & $11.1\,|\,13.9$ \\
     & Hard & $33.7\,|\,5.3$ & $73.6\,|\,94.6$ & $27.0\,|\,4.4$ & $35.2\,|\,82.2$ & $3.6\,|\,4.7$ & $4.0\,|\,5.6$ & $4.6\,|\,4.4$ & $2.2\,|\,2.0$ & $21.6\,|\,21.1$ & $629\,|\,691$ & $13.0\,|\,14.0$ \\
    \midrule
    \multicolumn{13}{l}{\textbf{\textit{Non-reasoning Models}}} \\
    \multirow{3}{*}{GLM-4.6V} & Easy & $31.0\,|\,2.4$ & $73.5\,|\,95.0$ & $32.0\,|\,4.1$ & $35.1\,|\,89.8$ & $3.9\,|\,3.9$ & $4.0\,|\,3.8$ & $2.7\,|\,2.5$ & $0.7\,|\,0.6$ & $11.5\,|\,9.9$ & $447\,|\,715$ & $4.1\,|\,4.1$ \\
     & Med & $23.0\,|\,1.0$ & $74.8\,|\,97.6$ & $26.0\,|\,1.1$ & $34.7\,|\,96.1$ & $3.6\,|\,3.4$ & $4.0\,|\,3.5$ & $2.7\,|\,2.5$ & $0.7\,|\,0.4$ & $11.4\,|\,8.6$ & $274\,|\,901$ & $2.7\,|\,6.0$ \\
     & Hard & $28.8\,|\,1.8$ & $74.7\,|\,98.7$ & $21.2\,|\,1.0$ & $38.4\,|\,96.1$ & $3.7\,|\,3.8$ & $4.0\,|\,3.7$ & $2.7\,|\,2.5$ & $0.6\,|\,0.5$ & $10.9\,|\,9.7$ & $289\,|\,582$ & $2.6\,|\,3.5$ \\
    \addlinespace[1.3pt]
    \multirow{3}{*}{DeepSeek-V3.2} & Easy & $39.3\,|\,16.0$ & $69.8\,|\,83.3$ & $29.0\,|\,16.0$ & $33.8\,|\,65.9$ & $4.0\,|\,2.8$ & $4.1\,|\,3.3$ & $4.7\,|\,5.0$ & $1.8\,|\,1.8$ & $17.0\,|\,16.4$ & $162\,|\,220$ & $2.4\,|\,2.8$ \\
     & Med & $24.0\,|\,7.0$ & $75.7\,|\,93.9$ & $23.0\,|\,7.0$ & $36.1\,|\,88.8$ & $4.1\,|\,2.8$ & $4.1\,|\,3.2$ & $4.6\,|\,4.9$ & $1.6\,|\,1.9$ & $16.3\,|\,16.0$ & $158\,|\,225$ & $2.2\,|\,2.7$ \\
     & Hard & $32.0\,|\,6.3$ & $72.9\,|\,95.9$ & $21.0\,|\,5.0$ & $38.1\,|\,90.4$ & $4.0\,|\,2.6$ & $4.1\,|\,3.2$ & $4.6\,|\,4.9$ & $1.7\,|\,1.9$ & $17.2\,|\,16.2$ & $138\,|\,206$ & $2.0\,|\,2.6$ \\
    \midrule
    \multicolumn{13}{l}{\textbf{\textit{Proprietary Models}}} \\
    \multirow{3}{*}{Gemini-2.5-Flash} & Easy & $32.0\,|\,23.0$ & $68.0\,|\,80.5$ & $25.5\,|\,16.1$ & $39.9\,|\,61.7$ & $3.3\,|\,2.4$ & $3.8\,|\,3.3$ & $3.4\,|\,4.1$ & $1.2\,|\,1.8$ & $13.8\,|\,16.0$ & {\color{gray}$109\,|\,136$} & {\color{gray}$1.5\,|\,1.7$} \\
     & Med & $29.0\,|\,4.0$ & $73.4\,|\,95.1$ & $24.2\,|\,7.3$ & $37.1\,|\,87.2$ & $3.1\,|\,2.3$ & $4.0\,|\,3.2$ & $4.6\,|\,4.3$ & $2.1\,|\,1.8$ & $19.7\,|\,15.6$ & {\color{gray}$102\,|\,146$} & {\color{gray}$1.8\,|\,1.7$} \\
     & Hard & $29.0\,|\,5.0$ & $72.5\,|\,94.4$ & $20.8\,|\,4.2$ & $39.3\,|\,88.3$ & $3.0\,|\,2.2$ & $4.0\,|\,3.1$ & $4.6\,|\,4.1$ & $2.1\,|\,2.1$ & $19.2\,|\,16.1$ & {\color{gray}$98\,|\,133$} & {\color{gray}$1.6\,|\,1.7$} \\
    \addlinespace[1.3pt]
    \multirow{3}{*}{\shortstack[l]{Gemini-3.1-Pro-\\Preview}} & Easy & $80.0\,|\,65.0$ & $57.8\,|\,52.1$ & $41.0\,|\,38.5$ & $30.4\,|\,44.5$ & $2.8\,|\,1.2$ & $3.6\,|\,2.1$ & $4.1\,|\,3.9$ & $1.5\,|\,1.5$ & $13.7\,|\,9.6$ & {\color{gray}$66\,|\,101$} & {\color{gray}$0.8\,|\,0.7$} \\
     & Med & $40.0\,|\,21.0$ & $70.6\,|\,84.4$ & $32.7\,|\,14.3$ & $36.1\,|\,79.9$ & $2.9\,|\,1.2$ & $3.7\,|\,2.2$ & $4.2\,|\,4.3$ & $1.7\,|\,1.6$ & $15.3\,|\,11.5$ & {\color{gray}$52\,|\,104$} & {\color{gray}$0.6\,|\,0.9$} \\
     & Hard & $33.0\,|\,19.0$ & $71.2\,|\,84.7$ & $31.9\,|\,13.3$ & $37.1\,|\,81.2$ & $3.1\,|\,1.1$ & $3.7\,|\,2.0$ & $4.0\,|\,4.3$ & $1.5\,|\,1.7$ & $14.3\,|\,11.1$ & {\color{gray}$59\,|\,86$} & {\color{gray}$0.7\,|\,0.7$} \\
    \midrule
    \multicolumn{10}{l}{\textbf{\textit{Search Baselines}}} & \multicolumn{1}{c}{\textit{TNC(k)}} & & \multicolumn{1}{c}{\textit{Budget}} \\
    \multirow{3}{*}{\shortstack[l]{MCTS/UCT\\{\scriptsize (Smargo)}}} & Easy & $33.0\,|\,5.0$ & $72.7\,|\,96.7$ & $30.0\,|\,0.0$ & $34.4\,|\,42.7$ & $4.9\,|\,3.1$ & $3.7\,|\,35.2$ & $5.3\,|\,4.8$ & $45.9\,|\,47.1$ & $7.2\,|\,7.8$ & $\text{--}\,|\,\text{--}$ & $200^\dagger\,|\,200^\dagger$ \\
     & Med & $32.0\,|\,1.0$ & $73.5\,|\,97.5$ & $26.0\,|\,0.0$ & $29.8\,|\,35.6$ & $5.7\,|\,2.5$ & $3.8\,|\,38.0$ & $5.2\,|\,4.8$ & $52.0\,|\,54.6$ & $8.7\,|\,9.1$ & $\text{--}\,|\,\text{--}$ & $200^\dagger\,|\,200^\dagger$ \\
     & Hard & $22.0\,|\,1.0$ & $74.7\,|\,98.0$ & $21.0\,|\,0.0$ & $40.6\,|\,37.8$ & $5.1\,|\,2.8$ & $3.8\,|\,49.6$ & $5.3\,|\,4.9$ & $63.1\,|\,65.3$ & $11.0\,|\,11.4$ & $\text{--}\,|\,\text{--}$ & $200^\dagger\,|\,200^\dagger$ \\
    \addlinespace[1.3pt]
    \multirow{3}{*}{KataGo-b18} & Easy & $97.0\,|\,49.0$ & $4.2\,|\,54.0$ & $97.0\,|\,53.0$ & $0.7\,|\,27.9$ & $1.5\,|\,6.8$ & $2.1\,|\,8.4$ & $\text{--}\,|\,\text{--}$ & $\text{--}\,|\,\text{--}$ & $\text{--}\,|\,\text{--}$ & $\text{--}\,|\,\text{--}$ & $200^\dagger\,|\,200^\dagger$ \\
     & Med & $75.0\,|\,46.0$ & $25.5\,|\,59.6$ & $62.0\,|\,44.0$ & $9.5\,|\,28.2$ & $5.0\,|\,8.7$ & $2.6\,|\,9.9$ & $\text{--}\,|\,\text{--}$ & $\text{--}\,|\,\text{--}$ & $\text{--}\,|\,\text{--}$ & $\text{--}\,|\,\text{--}$ & $200^\dagger\,|\,200^\dagger$ \\
     & Hard & $58.0\,|\,43.0$ & $48.1\,|\,67.0$ & $48.0\,|\,36.0$ & $16.8\,|\,29.0$ & $7.5\,|\,12.0$ & $2.9\,|\,10.2$ & $\text{--}\,|\,\text{--}$ & $\text{--}\,|\,\text{--}$ & $\text{--}\,|\,\text{--}$ & $\text{--}\,|\,\text{--}$ & $200^\dagger\,|\,200^\dagger$ \\
    \bottomrule
  \end{tabular}%
  }
\end{table*}

\begin{table*}[p]
  \centering
  \caption{Per-modality LLM accuracy results. Accuracy columns report first-move hit rate by input modality: Sym=Symbolic, Grd=Grid, S+G=Symbolic+Grid, Vis=Visual, and S+V=Symbolic+Visual. Trace-level diagnostics are reported separately in Table~\ref{tab:appendix_complete_results}.}
  \label{tab:appendix_full_llm_results}
  \scriptsize
  \setlength{\tabcolsep}{4pt}
  \renewcommand{\arraystretch}{0.88}
  \begin{tabular}{ll c ccccc}
    \toprule
    & & & \multicolumn{5}{c}{Accuracy (\%)} \\
    \cmidrule(lr){4-8}
    Model & Cond. & Diff. & Sym & Grd & S+G & Vis & S+V \\
    \midrule
    \multirow{6}{*}{Kimi-K2.5}
      & \multirow{3}{*}{$K=4$}
        & Easy & 57 & 47 & 49 & 51 & 56 \\
      & & Med  & 25 & 30 & 30 & 28 & 29 \\
      & & Hard & 33 & 25 & 38 & 37 & 37 \\
      \cmidrule(l){2-8}
      & \multirow{3}{*}{$K=\mathrm{None}$}
        & Easy & 29 & 20 & 35 & 22 & 33 \\
      & & Med  & 10 & 11 & 12 & 9 & 13 \\
      & & Hard & 9 & 1 & 4 & 3 & 5 \\
    \midrule
    \multirow{6}{*}{Qwen3-VL-235B}
      & \multirow{3}{*}{$K=4$}
        & Easy & 46 & 47 & 41 & 29 & 37 \\
      & & Med  & 23 & 27 & 27 & 26 & 22 \\
      & & Hard & 31 & 26 & 31 & 39 & 33 \\
      \cmidrule(l){2-8}
      & \multirow{3}{*}{$K=\mathrm{None}$}
        & Easy & 17 & 12 & 20 & 6 & 24 \\
      & & Med  & 11 & 9 & 7 & 6 & 11 \\
      & & Hard & 6 & 5 & 11 & 6 & 8 \\
    \midrule
    \multirow{6}{*}{GLM-4.6V}
      & \multirow{3}{*}{$K=4$}
        & Easy & 33 & 23 & 29 & 36 & 34 \\
      & & Med  & 17 & 23 & 28 & 22 & 25 \\
      & & Hard & 24 & 28 & 26 & 34 & 32 \\
      \cmidrule(l){2-8}
      & \multirow{3}{*}{$K=\mathrm{None}$}
        & Easy & 4 & 3 & 1 & 0 & 4 \\
      & & Med  & 0 & 3 & 1 & 0 & 1 \\
      & & Hard & 3 & 2 & 2 & 0 & 2 \\
    \midrule
    \multirow{6}{*}{Qwen3-VL-30B}
      & \multirow{3}{*}{$K=4$}
        & Easy & 34 & 32 & 35 & 31 & 39 \\
      & & Med  & 24 & 24 & 19 & 22 & 26 \\
      & & Hard & 36 & 30 & 31 & 31 & 23 \\
      \cmidrule(l){2-8}
      & \multirow{3}{*}{$K=\mathrm{None}$}
        & Easy & 9 & 4 & 15 & 4 & 18 \\
      & & Med  & 7 & 2 & 5 & 4 & 5 \\
      & & Hard & 6 & 2 & 2 & 1 & 7 \\
    \midrule
    \multirow{6}{*}{MiniMax-M2.5}
      & \multirow{3}{*}{$K=4$}
        & Easy & 41 & 36 & 36 & --- & --- \\
      & & Med  & 22 & 21 & 23 & --- & --- \\
      & & Hard & 25 & 27 & 27 & --- & --- \\
      \cmidrule(l){2-8}
      & \multirow{3}{*}{$K=\mathrm{None}$}
        & Easy & 10 & 12 & 18 & --- & --- \\
      & & Med  & 9 & 4 & 9 & --- & --- \\
      & & Hard & 3 & 8 & 8 & --- & --- \\
    \midrule
    \multirow{6}{*}{DeepSeek-R1-0528}
      & \multirow{3}{*}{$K=4$}
        & Easy & 32 & 27 & 38 & --- & --- \\
      & & Med  & 31 & 26 & 22 & --- & --- \\
      & & Hard & 28 & 41 & 32 & --- & --- \\
      \cmidrule(l){2-8}
      & \multirow{3}{*}{$K=\mathrm{None}$}
        & Easy & 13 & 18 & 20 & --- & --- \\
      & & Med  & 8 & 11 & 6 & --- & --- \\
      & & Hard & 4 & 4 & 8 & --- & --- \\
    \midrule
    \multirow{6}{*}{Qwen3-30B}
      & \multirow{3}{*}{$K=4$}
        & Easy & 20 & 37 & 23 & --- & --- \\
      & & Med  & 21 & 24 & 19 & --- & --- \\
      & & Hard & 24 & 39 & 30 & --- & --- \\
      \cmidrule(l){2-8}
      & \multirow{3}{*}{$K=\mathrm{None}$}
        & Easy & 15 & 9 & 17 & --- & --- \\
      & & Med  & 8 & 6 & 7 & --- & --- \\
      & & Hard & 3 & 3 & 5 & --- & --- \\
    \midrule
    \multirow{6}{*}{DeepSeek-V3.2}
      & \multirow{3}{*}{$K=4$}
        & Easy & 33 & 41 & 44 & --- & --- \\
      & & Med  & 21 & 29 & 22 & --- & --- \\
      & & Hard & 37 & 31 & 28 & --- & --- \\
      \cmidrule(l){2-8}
      & \multirow{3}{*}{$K=\mathrm{None}$}
        & Easy & 19 & 10 & 19 & --- & --- \\
      & & Med  & 6 & 8 & 7 & --- & --- \\
      & & Hard & 5 & 6 & 8 & --- & --- \\
    \bottomrule
  \end{tabular}
\end{table*}

\subsection{Metric Discriminative Power}
\label{app:metric_power}

Table~\ref{tab:metric_power} reports two complementary measures of metric discriminative power.  Cohen's $d$ is computed on correct vs.\ incorrect samples, while Spearman $\rho$ is computed across model-level metric means vs.\ accuracy.

\begin{table}[t]
  \centering
  \caption{Discriminative power of search-trace metrics.}
  \label{tab:metric_power}
  \small
  \setlength{\tabcolsep}{4pt}
  \begin{tabular}{lclcl}
    \toprule
    Metric & $d$ & Direction & $\rho$ & Assessment \\
    \midrule
    $\swr$ & 2.07 & corr.\ $<$ wrong & $-0.82$ & Strong \\
    $\sfh$ & 1.02 & corr.\ $>$ wrong & $+0.71$ & Strong \\
    $\scr$ & 0.57 & corr.\ $<$ wrong & $-0.21$ & Moderate \\
    $\tmd$ & 0.11 & $\approx$ & $+0.82$ & Cross-model \\
    $\tmf$ & 0.46 & corr.\ $>$ wrong & $+0.75$ & Cross-model \\
    $\itt$ & 0.28 & corr.\ $>$ wrong & $+0.46$ & Moderate \\
    $\tnc$ & 0.05 & $\approx$ & $+0.54$ & Moderate \\
    \bottomrule
  \end{tabular}
\end{table}

\begin{figure*}[t]
  \centering
  \includegraphics[width=.95\textwidth]{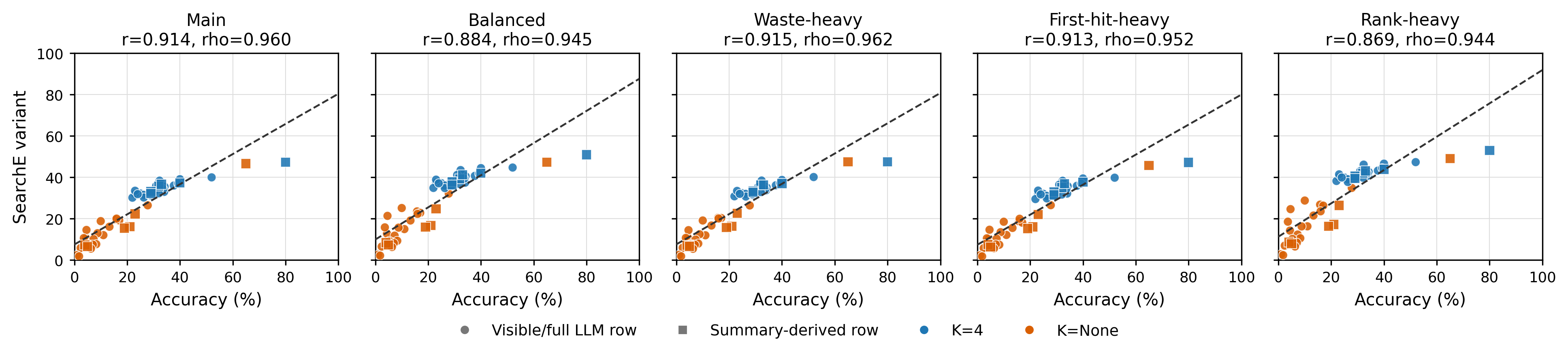}
  \caption{Accuracy--SearchE scatter plots under alternative SearchE weight schemes.  The relationship remains monotonic across nearby weight choices, supporting the use of SearchE as a compact search-organization summary rather than a fragile fitted score.}
  \label{fig:weight_sensitivity}
\end{figure*}

\paragraph{Weight sensitivity.}
We recomputed SearchE under five interpretable weight schemes over the non-baseline LLM rows used for the main correlation analysis.  Table~\ref{tab:weight_sensitivity} reports the correlation between each SearchE variant and accuracy; Figure~\ref{fig:weight_sensitivity} shows the corresponding scatter plots.  The main $(0.5,0.3,0.2)$ setting is not uniquely tuned to the data: nearby balanced, waste-heavy, first-hit-heavy, and rank-heavy variants preserve high Pearson and rank correlation.  This supports using SearchE as a compact summary of search organization rather than a fragile fitted score.

\begin{table}[t]
  \centering
  \caption{Weight sensitivity of SearchE variants against accuracy on non-baseline LLM rows.  Weights are ordered as wrong-branch waste, first-hit behavior, and correct-rank cost.}
  \label{tab:weight_sensitivity}
  \small
  \setlength{\tabcolsep}{3pt}
  \begin{tabular}{lccc}
    \toprule
    Scheme & Weights & Pearson $r$ & Spearman $\rho$ \\
    \midrule
    Main & .50/.30/.20 & .914 & .960 \\
    Balanced & .33/.33/.33 & .884 & .945 \\
    Waste-heavy & .60/.20/.20 & .915 & .962 \\
    First-hit-heavy & .40/.40/.20 & .913 & .952 \\
    Rank-heavy & .40/.20/.40 & .869 & .944 \\
    \bottomrule
  \end{tabular}
\end{table}

\section{Evaluation Settings}
\label{app:experiment_settings}

All LLM evaluations use the same task interface: bounded-candidate runs provide $K=4$ options, while open-search runs use $K=\mathrm{None}$ and require the model to generate the first move.  Final answers are scored after coordinate normalization.  Decoding settings vary by model family and provider; most long-form runs use temperature $0.7$ with a large output budget, while some self-hosted or deterministic fill-in runs use temperature $0$--$0.5$.

For reproducibility, each evaluation record links the normalized problem state, difficulty, modality, candidate set, side to play, model family, decoding setting, output budget, and final normalized answer.  This is sufficient to recompute answer metrics from final moves, while process metrics are interpreted only for the visible reasoning exposed by the model.

\begin{table*}[t]
  \centering
  \caption{Reproducibility summary for model evaluations.  Provider names follow the main experimental setup; temperature and output budgets summarize the evaluated run families.}
  \label{tab:api_settings}
  \small
  \setlength{\tabcolsep}{3pt}
  \begin{tabular}{lllll}
    \toprule
    Model family & Provider & Modalities & Temp. & Max output \\
    \midrule
    Kimi-K2.5 & Moonshot AI & S/G/S+G & 0.7 & 32k \\
    Qwen3-VL-235B-Thinking & Alibaba/Qwen & S/G/V/S+V/S+G & 0.7 & 32k \\
    Qwen3-VL-30B-Thinking & self-hosted vLLM & S/G/V/S+V/S+G & 0.5--0.7 & 30k--32k \\
    MiniMax-M2.5 & MiniMax & S/G/S+G & 0.7 & 32k \\
    DeepSeek-R1-0528 & DeepSeek & S/G/S+G & 0.7 & 32k \\
    DeepSeek-V3.2 & DeepSeek & S/G/S+G & 0.7 & 32k \\
    GLM-4.6V & Zhipu AI & S/G/V/S+V/S+G & 0.7 & 32k \\
    Gemini-2.5 family & Google & S/G/V/S+V/S+G & 0--default & -- \\
    Gemini-3 family & Google & S/G/V/S+V/S+G & 0--default & -- \\
    \bottomrule
  \end{tabular}
\end{table*}

For proprietary systems that expose only reasoning summaries, tree size and token-scale measurements are treated as summary-derived observations.  They are useful for comparing visible output organization, but they are not claims about hidden internal computation.

\section{MCTS/KataGo Settings}
\label{app:nonllm_settings}

We compare LLM traces with two non-LLM search references: Smargo MCTS/UCT and KataGo.  These are not token-equivalent baselines.  They provide search-reference runs that show how non-language search procedures allocate fixed playout or visit budgets on the same normalized tsumego board states.

\paragraph{Root modes.}
We use two root settings.  \emph{Open-root} starts from the original board and lets the search procedure select among legal root moves.  For Smargo this uses its local legal-move generator; for KataGo this means no root candidate restriction in the analysis query.  \emph{Restricted-root} uses the benchmark $K=4$ interface: only the four listed answer options are permitted at root after filtering occupied or invalid coordinates.  Open-root runs test whether final distractors are naturally search-visible, while restricted-root runs align the search references with the LLM four-choice interface.

\paragraph{Smargo MCTS/UCT.}
Smargo is used as an unguided MCTS/UCT reference, using the public implementation by Sun-Yize et al.  Each problem receives 200 MCTS playouts with fixed seeds.  In the restricted-root setting, the root action set is limited to the four benchmark candidates; in the open-root setting, the search may choose among legal local root moves.  The candidate-construction audit uses the early part of the same playout trajectory to check whether selected distractors are encountered under shallow search.

Smargo selection uses
\begin{equation}
  Q(s,a) + c_{\mathrm{puct}} P(s,a)\frac{\sqrt{N(s)}}{1+N(s,a)},
\end{equation}
with $c_{\mathrm{puct}}=5.0$ in our runs.  Root priors are uniform over the permitted root moves.  At a newly expanded leaf, Smargo expands the available local moves with uniform priors, then performs a rollout until a local tsumego terminal condition is reached, no valid continuation remains, or the board-area depth limit is reached.  The terminal condition is local to the life-and-death solver rather than full-game territory scoring, so the baseline is best interpreted as a local search reference rather than a general Go-playing agent.

\paragraph{KataGo.}
KataGo is used as a neural-guided search reference.  We use KataGo v1.16.4 with the b18 network and a 200-visit budget per problem under Chinese rules, komi 7.5, and a $19\times19$ board.  The analysis configuration uses deterministic root behavior with resignation disabled, no root noise, and principal-variation visit reporting enabled.  In restricted-root runs, the root is limited to the benchmark's four candidates; in open-root runs, no such candidate restriction is imposed.  The same visit trajectory is used for shallow-budget audit at visits 1--30.

\paragraph{Input and output mapping.}
Both references consume the normalized board state, not natural-language puzzle text.  The board matrix is converted to the shared Go coordinate system \texttt{A--T} with \texttt{I} skipped.  Smargo internally indexes board points and maps them back to benchmark coordinates; KataGo receives the same normalized stones and side-to-play information through its analysis interface.  Top-1 accuracy is computed from the final root move with the largest visit count after coordinate normalization.  Illegal, occupied, pass, missing, or off-format outputs are treated as non-correct; in restricted-root runs occupied options are filtered before evaluation.

For metric mapping, Smargo exposes an explicit final MCTS tree, so $\tnc(k)$ is the number of reachable tree nodes after $k$ playouts, with final $\tnc$ reported at $k=200$.  KataGo exposes root statistics and principal variations rather than a complete internal search tree, so KataGo tree-size columns are not interpreted as internal $\tnc$.  For both references, root-level SearchE components are computed from the root visit distributions: $\swr=1-$correct root visit share, $\sfh$ indicates whether the earliest top root move is correct, $\scr$ is derived from the final correct-candidate rank and root branch count, and $\ssc$ counts top-root switches over the 1--200 trajectory.  Budget is reported as playouts for Smargo and visits for KataGo, never as language tokens.

\begin{table*}[t]
  \centering
  \caption{Reproducible non-LLM search-reference settings.}
  \label{tab:mcts_katago_settings}
  \small
  \setlength{\tabcolsep}{5pt}
  \begin{tabular}{llllll}
    \toprule
    Method & Search type & Root mode & Budget & N & Output used \\
    \midrule
    Smargo MCTS/UCT & unguided MCTS/UCT & $K=4$ & 200 playouts & 300 & root distribution + full tree \\
    Smargo MCTS/UCT & unguided MCTS/UCT & open & 200 playouts & 300 & root distribution + full tree \\
    KataGo b18 & neural-guided search & $K=4$ & 200 visits & 300 & root distribution + principal variations \\
    KataGo b18 & neural-guided search & open & 200 visits & 300 & root distribution + principal variations \\
    \bottomrule
  \end{tabular}
\end{table*}

\begin{figure}[t]
  \centering
  \includegraphics[width=\columnwidth]{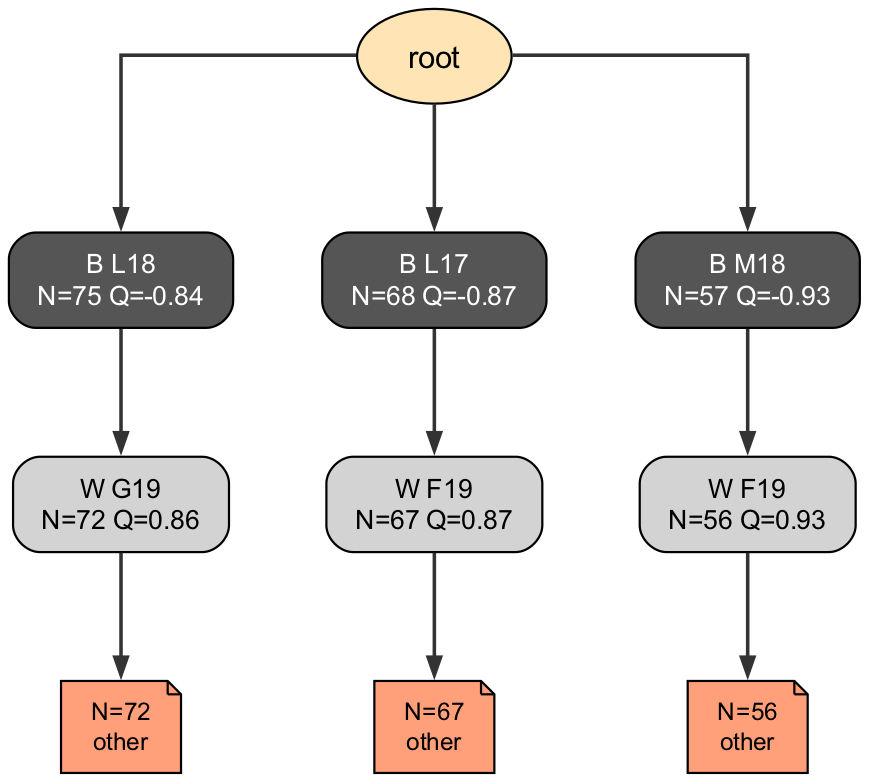}
  \caption{Representative Smargo MCTS/UCT tree for a restricted-root $K=4$ problem after 200 playouts.  The same trajectory format is used to compute MCTS tree-size diagnostics such as $\tnc(200)$, maximum depth, and maximum branching factor.}
  \label{fig:smargo_tree_example}
\end{figure}

Figure~\ref{fig:smargo_tree_example} shows a representative final Smargo tree for a restricted-root $K=4$ problem after 200 playouts.  The example is included to make the reported $\tnc$, depth, and branching-factor quantities concrete: unlike KataGo, Smargo exposes the explicit explored tree, so each node-count statistic is computed from the saved MCTS tree rather than inferred from principal variations.

Search-reference budgets are kept in their native units: playouts for Smargo, visits for KataGo, and tokens for LLMs.  The main non-LLM setting uses 200 playouts or visits per problem; smaller intermediate budgets are retained only for trajectory analysis and are not used as separate evaluation metrics.

\section{Prompt Templates}
\label{app:prompt_templates}

We include faithful templates of the Chinese task prompts used for evaluated models.  The extraction prompt is already described in Appendix~\ref{app:extraction_validation}; here we report only the two task prompts used for bounded candidate discrimination and open first-move generation.

\paragraph{Bounded candidate template.}
\begin{quote}\small
You are a top-level Go master.  Please carefully analyze this Go life-and-death problem, reason deeply, and choose the optimal first move for Black from the given options.  The input then provides one or more board presentations: coordinate lists, a numeric board matrix, a rendered board image, or their combinations.  The prompt lists the candidate moves as \texttt{A. <coord>}, \texttt{B. <coord>}, \ldots.  Please think step by step about the continuations after each option, and finally state your answer clearly.  Answer format: at the end, explicitly write ``answer is X'', where X is one of the option labels.
\end{quote}

\paragraph{Open-search template.}
\begin{quote}\small
You are a top-level Go master.  Please carefully analyze this Go life-and-death problem, reason deeply, and find the optimal first move for the side to play.  The input then provides one or more board presentations: coordinate lists, a numeric board matrix, a rendered board image, or their combinations.  No candidate moves are provided.  Please think step by step about possible moves and their continuations, and finally state your answer clearly.  Answer format: at the end, explicitly write ``answer is X'', where X is a board coordinate such as \texttt{T18}.
\end{quote}

\paragraph{Example bounded response.}
The following illustrates the expected response style: free-form reasoning followed by an explicit final option label.
\begin{quote}\small
Option A plays directly at the vital point and leaves White without enough eye space after the local reply.  Option B lets White connect and keep liberties, so Black loses the forcing sequence.  Options C and D are farther from the eye-shape weakness and do not solve the local life-and-death problem.  Therefore the strongest first move is option A.  \textbf{Answer is A.}
\end{quote}

\section{Example Responses and Extracted Trees}
\label{app:response_tree_examples}

We report eight representative audited process-tree examples from the bounded $K=4$ candidate interface.  The cases are labeled as \texttt{K-4-E-<id>--<model>} and are chosen to place one model-family example in each panel of a single-page montage.  Each panel shows the final extracted tree for one model response, together with the selected option and verified answer.

The resulting cases expose several recurring patterns.  Correct traces may be short when the key first move is introduced immediately, or longer when the model explicitly verifies several local replies before choosing.  Wrong traces are often still structured: they mention the correct region but either attach the terminal judgment to the wrong branch, drift in coordinate reconstruction, or accept a local eye-shape claim that does not match the verified answer.  This is why the process tree is useful beyond final accuracy: it records where the visible search path loses alignment with the answer key.

\begin{table}[t]
  \centering
  \caption{Displayed $K=4$ extracted-tree cases.  The case labels follow \texttt{K-4-E-<id>--<model>}.}
  \label{tab:tree_case_json_display}
  \scriptsize
  \setlength{\tabcolsep}{2pt}
  \begin{tabular}{lllll}
    \toprule
    Case & Tier & Model & Outcome & Correct \\
    \midrule
        K-4-E-001 & Easy & Qwen3-30B & wrong & T15 \\
    K-4-E-002 & Easy & Kimi-K2.5 & correct & P14 \\
    K-4-E-003 & Easy & MiniMax-M2.5 & wrong & T13 \\
    K-4-E-004 & Easy & DeepSeek-V3.2 & correct & P15 \\
    K-4-E-005 & Easy & DeepSeek-R1 & correct & T15 \\
    K-4-E-006 & Easy & GLM-4.6V & correct & S13 \\
    K-4-E-007 & Easy & Qwen3-VL & correct & S17 \\
    K-4-E-008 & Easy & Gemini-2.5-Pro & correct & T13 \\
    \bottomrule

  \end{tabular}
\end{table}

\begin{figure*}[t]
  \centering
  \includegraphics[width=.98\textwidth]{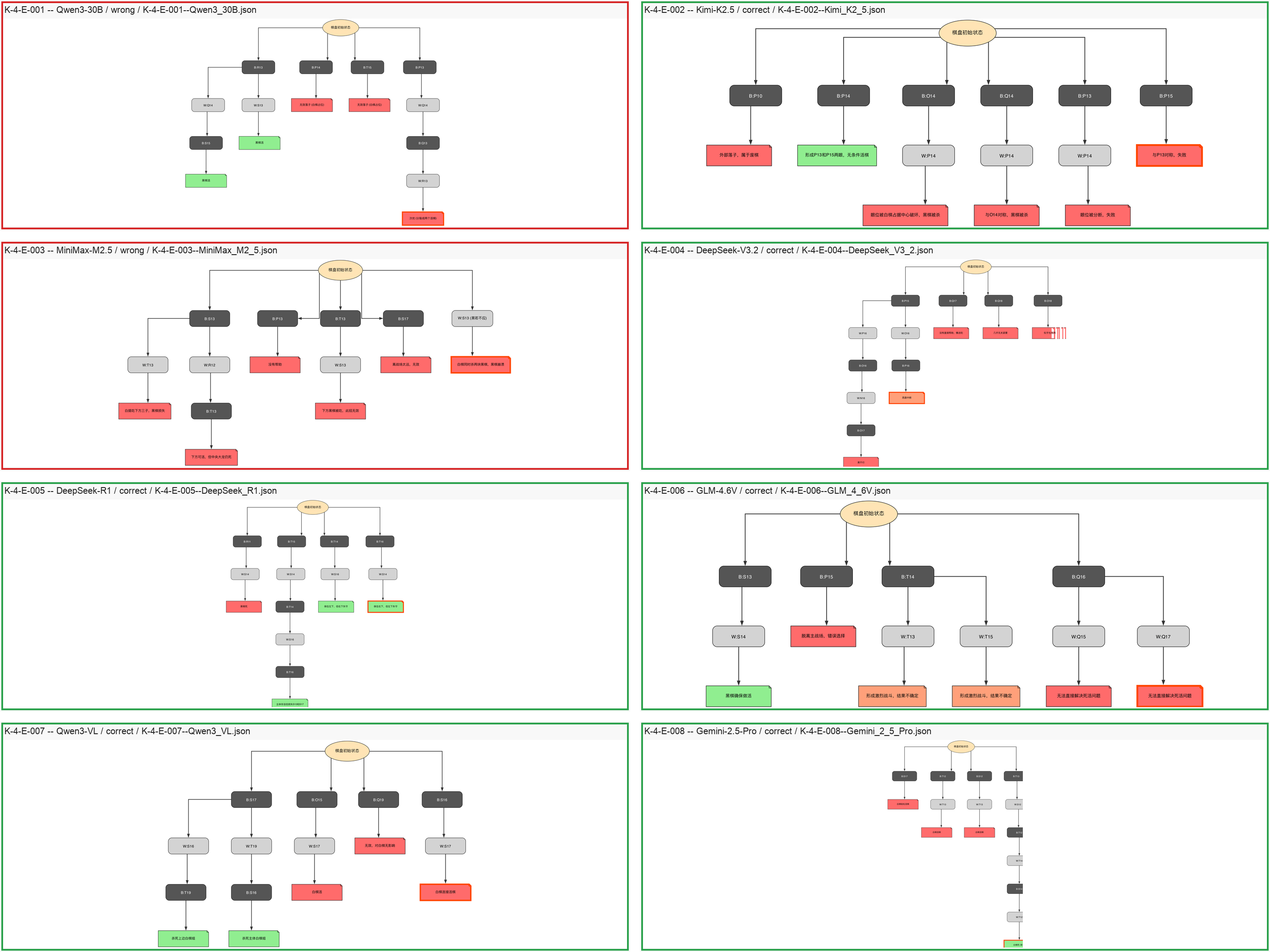}
  \caption{$K=4$ extracted process trees for eight displayed Easy examples.  Green panels denote correct final answers and red panels denote wrong final answers.}
  \label{fig:model_tree_montage_easy}
\end{figure*}

\setcounter{table}{16}
\setcounter{section}{9}

\section{Domain-Specialized Model: Logos}
\label{app:logos}

We do not include Logos in the quantitative tables because the available setup does not provide outputs under the same prompt and response format.  The qualitative distinction is still important: a Go-specialized policy model may output plausible whole-game moves while failing at local life-and-death search because the objective is not merely to continue the game, but to organize adversarial forcing lines around life, death, capture, and eye-shape constraints.

\begin{figure*}[t]
  \centering
  \includegraphics[width=.94\textwidth]{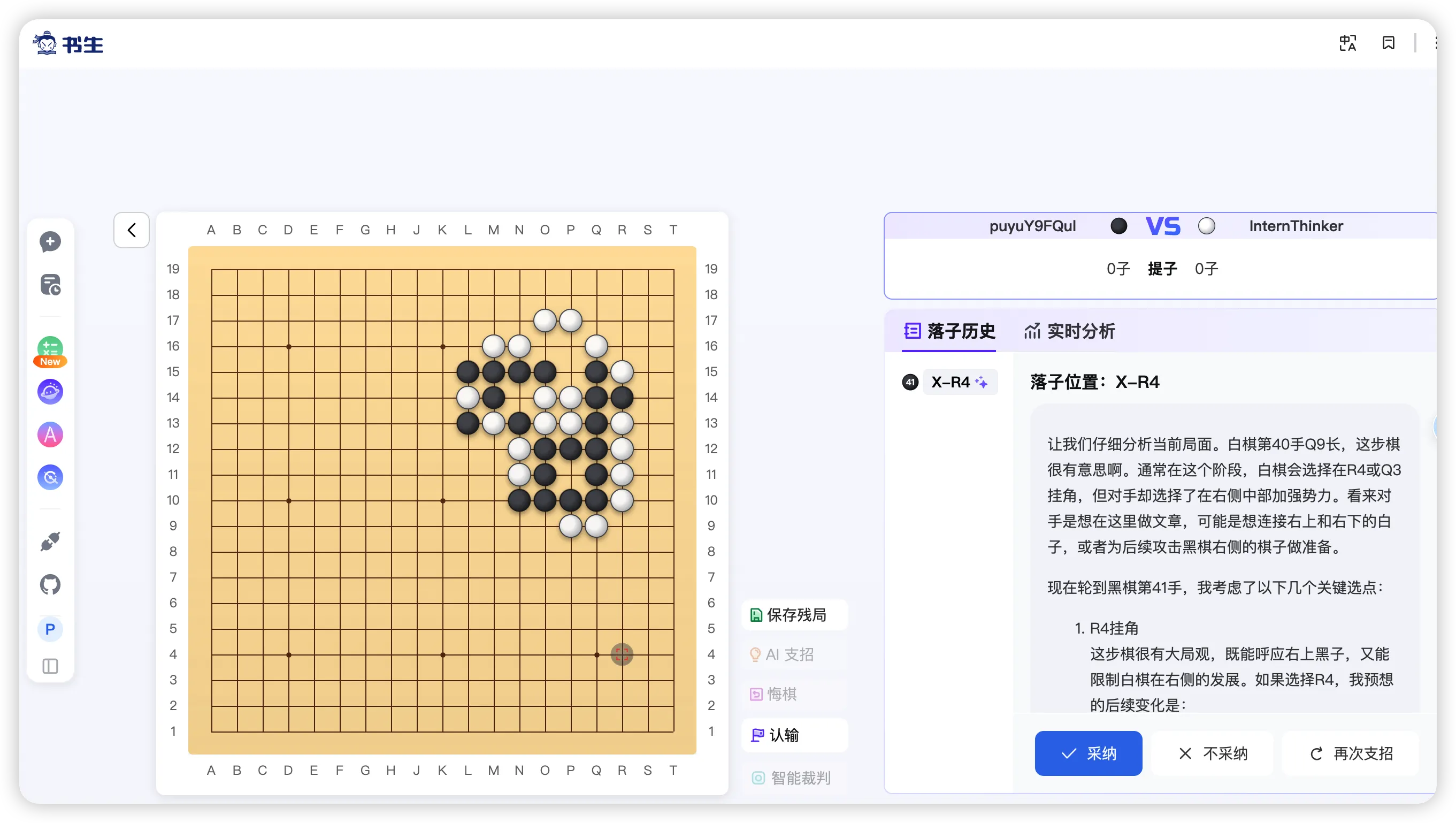}
  \caption{Qualitative Logos probe on the first benchmark problem.  The verified first move is \texttt{P15} or equivalent \texttt{N14}, whereas Logos returns \texttt{R4}.  The response behaves like a general Go continuation analysis rather than a local life-and-death solution trace, so it is treated as diagnostic evidence for response-format mismatch rather than as a scored benchmark run.}
  \label{fig:logos_probe}
\end{figure*}

Figure~\ref{fig:logos_probe} shows a manual probe using the first benchmark problem in the online Logos interface.  For this problem, the verified answer set is \texttt{P15} or the equivalent first move \texttt{N14}, as shown in Appendix~\ref{tab:normalized_example}.  Logos instead returns \texttt{R4}, and its explanation is framed as ordinary game continuation and positional commentary rather than as a local life-and-death solution.  It discusses global strength and follow-up direction, but does not identify the local live-or-kill objective, enumerate candidate first moves, or verify forcing replies.  This case is not used as a quantitative result; it illustrates why a domain-specialized Go interface is not automatically comparable to \tsumego{} unless its prompt, answer format, and trace extraction are aligned with local tsumego solving.

\section{Failure Analysis}
\label{app:failure_modes}

\begin{figure}[t]
  \centering
  \includegraphics[width=.96\columnwidth]{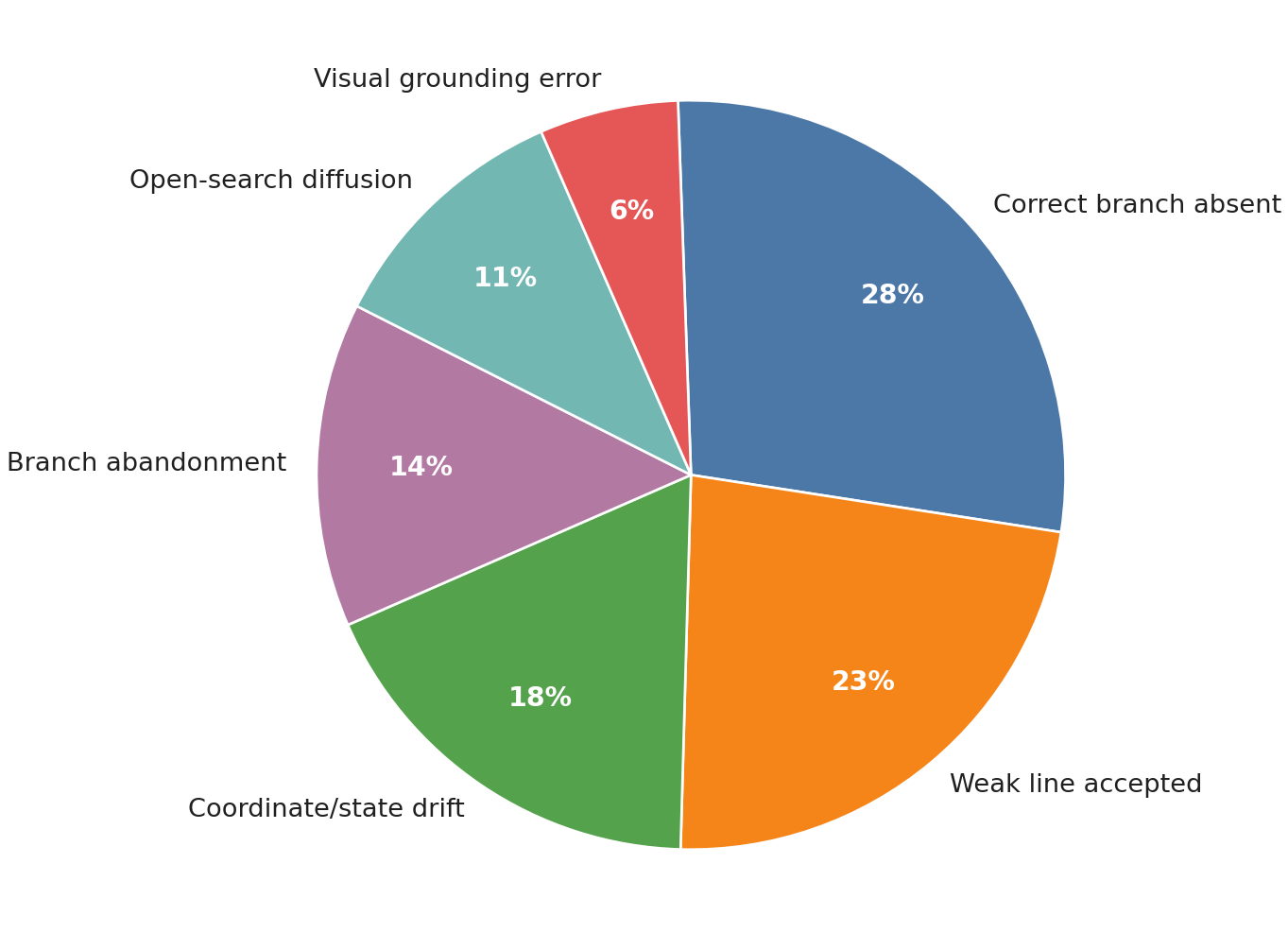}
  \caption{Failure-mode distribution from the balanced qualitative audit.  The dominant errors concern search organization and local verification rather than output formatting.}
  \label{fig:failure_modes}
\end{figure}

We audit wrong answers with a balanced slice across model families, difficulty tiers, and both $K=4$ and open-search interfaces.  Figure~\ref{fig:failure_modes} summarizes the dominant error sources.  The largest categories are search-organization failures rather than surface presentation errors: the solver either never enters the correct local branch, accepts a weak local line as decisive, or drifts from the intended coordinate and board state.

The distribution suggests a layered failure structure.  At the perceptual layer, visual grounding errors arise when the position is encoded with an incorrect stone, color, or occupancy relation; these errors are comparatively infrequent, but they poison all later tactical computation because the solver is reasoning over the wrong local state.  At the state-computation layer, coordinate/state drift reflects failures to maintain liberties, adjacency, captures, or coordinate identities across a multi-step line.  These cases are not simply notation mistakes: they indicate that the internal board update is no longer synchronized with the candidate sequence being evaluated.

At the search layer, correct-branch absence and open-search diffusion expose two complementary weaknesses.  In the bounded interface, the correct candidate may be available but not treated as the main tactical branch.  In the open interface, the model often expands many plausible-looking moves without converging to the vital point, which produces breadth without a decisive local objective.  Finally, weak-line acceptance and branch abandonment occur at the adversarial verification layer: the solver reaches a relevant region, but evaluates the opponent's reply too optimistically, prematurely terminates the line, or switches away from a still-viable forcing branch.  Thus the common failure is not lack of Go vocabulary, but failure to keep perception, board-state computation, candidate search, and adversarial verification aligned.

\section{Discussion and Limitations}
\label{app:discussion}

\paragraph{What \tsumego{} measures.}
\tsumego{} measures local adversarial search organization rather than general Go-playing strength.  Its main target is how a model plans a reasoning path and allocates observable reasoning resources across competing candidates: whether it enters the vital branch, sustains verification under opponent replies, avoids spending effort on shallow wrong lines, and reallocates effort when a line is refuted.  The process-tree view therefore separates the amount of reasoning from where that reasoning is placed, making branch selection, continuation depth, refutation checking, and backtracking visible as distinct planning behaviors.

\paragraph{Human--AI interaction context.}
This benchmark also exposes a mismatch between many current optimization settings and adversarial planning tasks.  Modern instruction tuning and deployment feedback often emphasize cooperative, non-adversarial, multi-turn interaction: the user can clarify intent, correct mistakes, narrow the search space, or accept partial progress.  That setting is valuable, but it gives the model external scaffolding for planning.  In \tsumego{}, the opponent does not cooperate, hidden correction is unavailable, and the model must maintain the local state while anticipating refutations.  The resulting failures therefore should not be read only as Go-specific errors; they indicate that models optimized for helpful interactive dialogue may still lack robust autonomous control over where reasoning effort is spent when the environment actively pushes back.

\paragraph{Trace observability.}
Search trees are extracted from observable CoT traces, so our process metrics describe the visible reasoning path rather than the model's latent computation.  This limitation is also an opportunity: future work could analyze internal activations, attention/state trajectories, or planner--verifier signals to recover latent search paths and compare them with the externalized trace.

\paragraph{Future work.}
A natural next step is to expand from local tsumego to broader agent-system settings where candidate generation, state tracking, tool use, memory, and adversarial feedback interact over longer horizons.  Another direction is to compare autonomous solving with human-guided multi-turn repair, measuring when interaction scaffolding compensates for weak internal search control.  Finally, \tsumego{} points to targeted training frameworks that supervise branch prioritization, opponent-response verification, state maintenance, backtracking, and search-budget allocation rather than only final answers or fluent rationales.

\end{document}